\documentclass{article}

\usepackage{arxiv}

\usepackage[utf8]{inputenc} 
\usepackage[T1]{fontenc}    
\usepackage{hyperref}       
\usepackage{url}            
\usepackage{booktabs}       
\usepackage{nicefrac}       
\usepackage{microtype}      
\usepackage{lipsum}
\usepackage{graphicx}
\graphicspath{ {./images/} }

\usepackage{amsmath,amsfonts}
\usepackage{algorithmic}
\usepackage{algorithm}
\usepackage{array}
\usepackage[caption=false,font=normalsize,labelfont=sf,textfont=sf]{subfig}
\usepackage{textcomp}
\usepackage{stfloats}
\usepackage{verbatim}
\usepackage{cite}
\usepackage{flushend}
\usepackage{amsthm}

\usepackage{epsfig}
\usepackage{tabularx}
\usepackage{color}

\usepackage{amsmath,amsfonts,bm}
\usepackage{xcolor}
\usepackage{fontawesome}

\def\eqref#1{(\ref{#1})}

\def\1{\bm{1}}

\def\rvh{{\mathbf{h}}}

\def\rvs{{\mathbf{s}}}

\def\rvz{{\mathbf{z}}}

\def\rmA{{\mathbf{A}}}

\def\rmE{{\mathbf{E}}}

\def\rmH{{\mathbf{H}}}
\def\rmI{{\mathbf{I}}}

\def\rmM{{\mathbf{M}}}

\def\rmX{{\mathbf{X}}}

\def\rmZ{{\mathbf{Z}}}

\DeclareMathAlphabet{\mathsfit}{\encodingdefault}{\sfdefault}{m}{sl}
\SetMathAlphabet{\mathsfit}{bold}{\encodingdefault}{\sfdefault}{bx}{n}

\def\gG{{\mathcal{G}}}

\def\gN{{\mathcal{N}}}

\newcommand{\R}{\mathbb{R}}

\newcommand{\T}{\textup{T}}

\title{GRAM: An Interpretable Approach for Graph Anomaly Detection using Gradient Attention Maps\\}

\author{
 Yifei Yang \\
  School of Electronic Information\\
  Wuhan University\\
  Wuhan, Hubei 430072 \\
  \texttt{yfyang@whu.edu.cn} \\
   \And
 Peng Wang \\
  Division of Natural and Applied Sciences\\
  Duke Kunshan University\\
  Kunshan, Jiangsu 215316 \\
  \texttt{pw140@duke.edu} \\
  \And
  \And
 Xiaofan He \\
  School of Electronic Information\\
  Wuhan University\\
  Wuhan, Hubei 430072 \\
  \texttt{xiaofanhe@whu.edu.cn} \\
   \And
 Dongmian Zou \\
  Division of Natural and Applied Sciences\\
  Duke Kunshan University\\
  Kunshan, Jiangsu 215316 \\
  \texttt{dongmian.zou@dukekunshan.edu.cn} \\
  \And
}

\begin{document}
\maketitle
\begin{abstract}
Detecting unusual patterns in graph data is a crucial task in data mining. However, existing methods face challenges in consistently achieving satisfactory performance and often lack interpretability, which hinders our understanding of anomaly detection decisions. In this paper, we propose a novel approach to graph anomaly detection that leverages the power of interpretability to enhance performance. Specifically, our method extracts an attention map derived from gradients of graph neural networks, which serves as a basis for scoring anomalies. Notably, our approach is flexible and can be used in various anomaly detection settings. In addition, we conduct theoretical analysis using synthetic data to validate our method and gain insights into its decision-making process. To demonstrate the effectiveness of our method, we extensively evaluate our approach against state-of-the-art graph anomaly detection techniques on real-world graph classification and wireless network datasets. The results consistently demonstrate the superior performance of our method compared to the baselines.
\end{abstract}


\section{Introduction}
\label{sec:intro}
\begin{figure}[t]
  \centering
  \subfloat[]{\includegraphics[width=0.65\columnwidth]{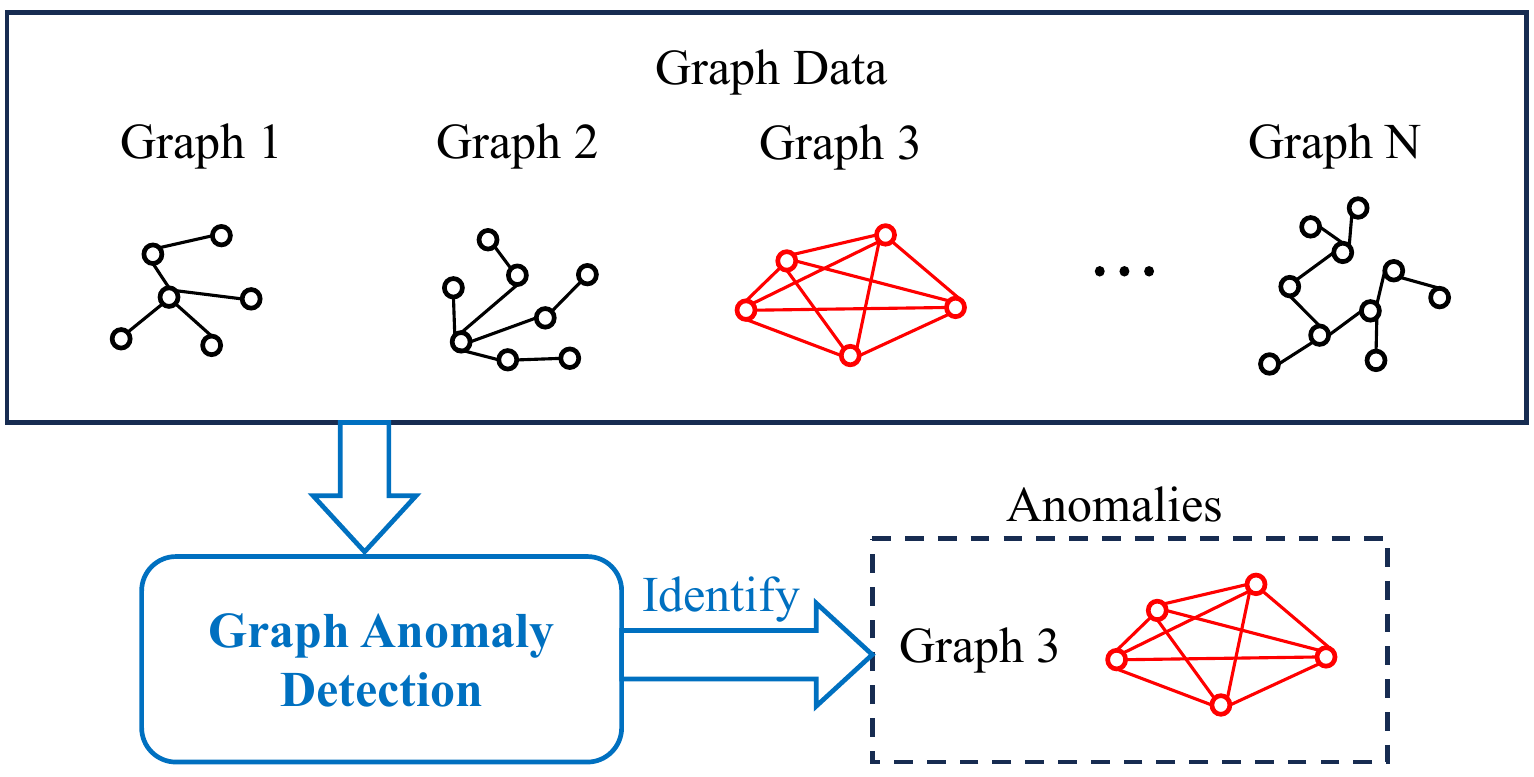}%
  \label{node-level}}
  \hfil
  \subfloat[]{\includegraphics[width=0.65\columnwidth]{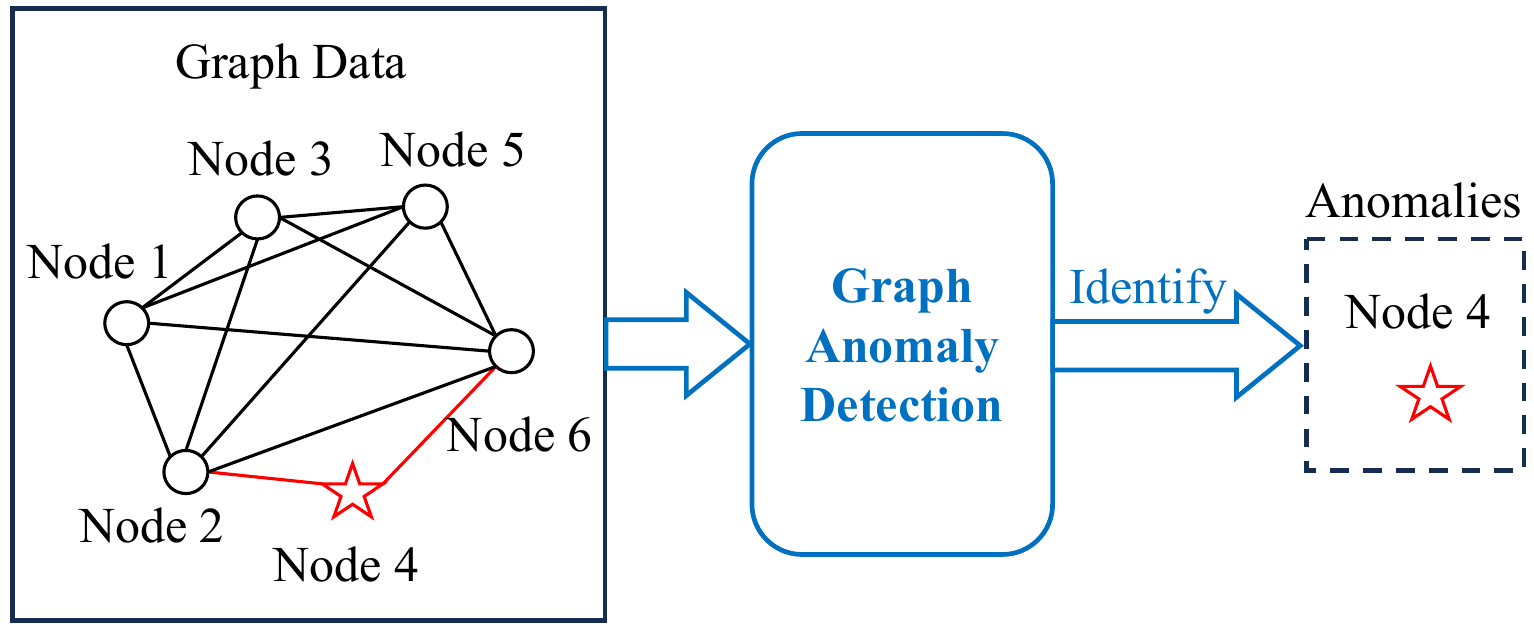}%
  \label{graph-level}}
  \caption{{Illustration of graph anomaly detection (GAD) tasks. (a) A graph is anomalous within a given dataset of graphs. (b) A node is anomalous within a given graph, where the graph data may comprise multiple graphs.}}
  \label{GAD}
\end{figure}

Anomalies refer to objects or instances that deviate significantly from the standard, normal, or expected behavior. They manifest across diverse domains, such as fake news in social media \cite{nguyen2020fang}, untrustworthy comments in online comment systems \cite{kumar2018rev2}, network intrusions \cite{miao2020attack}, telecom fraud \cite{wu2024beyond}, and financial fraud \cite{zhang2022efraudcom}. The study and detection of anomalies have become a crucial task, necessitating dedicated research efforts. Typically, the detection task involves identifying outlier data points in the feature space \cite{pang2021deep}, which often overlooks the relational information present in real-world data. Graphs, on the other hand, are frequently employed to represent structural and relational information, thereby posing the challenge of graph anomaly detection (GAD) \cite{akoglu2015graph}. GAD involves identifying anomalous graph elements such as nodes, edges, and subgraphs, within the dataset of a single graph or each individual graph within a collection of multiple graphs. Moreover, it extends to the task of detecting anomalous graphs within a graph set or database \cite{jie2019block}. Fig.~\ref{GAD} illustrates these two types of tasks. 

Traditional anomaly detection techniques face challenges in effectively addressing the GAD problem because of the complex relational dependencies and irregular structures associated with graph data. However, the recent advent of geometric deep learning and graph neural networks (GNNs) has generated significant interest in utilizing these advanced methods for GAD \cite{ma2021comprehensive}. Recent developments in GNNs have shown their effectiveness in capturing comprehensive information from graph structures and node attributes, making them widely used for GAD problems. However, these GAD methods are based on neural networks and are often regarded as black-box models. This lack of interpretability poses challenges in explaining how these models perform anomaly detection and make decisions. The inability to understand and interpret the results of anomaly detection may hinder end users, who rely on these results to make informed decisions. In addition, the effectiveness of GNN-based GAD methods may vary across different datasets, especially because of the complex relations and heterogeneity in graph data. 

To address these limitations, we propose a novel approach to GAD based on GNN interpretation in this work. Our method derives anomaly scores from gradient attention maps of GNNs, and we name it accordingly as ``GRAM'' (an abbreviation of ``\textbf{GR}adient \textbf{A}ttention \textbf{M}ap''). Our method is designed for the unsupervised GAD task where the underlying GNNs are trained solely on normal graph data. In this case, the trained GNNs only capture characteristics of normal graphs. Therefore, anomalous graph elements will exhibit distinctive highlighted regions when we examine the gradient maps of the entire graph representation with respect to node feature embeddings. GRAM does not rely on a specific GNN architecture. In the testing phase, it produces anomaly scores for both individual nodes and the entire graph, which serve as indicators of the likelihood of anomalies.

To validate the effectiveness of our proposed method, we conduct comprehensive evaluations encompassing both theoretical and empirical analyses. Firstly, we perform a theoretical analysis using synthetic data to illustrate the regime of GRAM and understand how it works. Subsequently, we conduct empirical evaluations and compare with state-of-the-art baseline methods using realistic graph classification and wireless network datasets to further validate the performance of GRAM. 

The remaining sections of the paper are organized as follows. Section \ref{sec:related_works} surveys related works. Section \ref{Proposed_method} presents the GRAM model and explains how to use it for unsupervised GAD. Section \ref{Analys_GRAM_VGAE} conducts theoretical analysis based on synthetic data. Section \ref{sec:experiments} demonstrates experiments to evaluate the performance of GRAM. Section \ref{sec:Conclusion} concludes the paper.

\section{Related Works}\label{sec:related_works}
Traditional anomaly detection methods, such as the Local Outlier Factor (LOF) \cite{breunig2000lof} and the One-Class Support Vector Machine (OCSVM) \cite{scholkopf2001estimating, erfani2016high}, have been widely utilized for detecting anomalies based on feature representations of data. With the rise of deep learning, researchers have turned to neural networks to extract more meaningful and complex features from high-dimensional data for anomaly detection \cite{zhai2016deep, zong2018deep, sabokrou2018adversarially, pidhorskyi2018generative, zenati2018adversarially, perera2019ocgan, xie2020unsupervised, lai2020robust, chen2020generative, zha2020meta, liguori2021adversarial, chen2022utrad, lai2023robust}. However, these methods are primarily designed for Euclidean data and often struggle to effectively leverage the structural information inherent in graphs, resulting in a suboptimal performance for GAD tasks. 

To this end, many recent anomaly detection methods specifically tailored for graph data have been proposed. By leveraging GNNs, these methods can better model the relationships and dependencies among nodes and edges in a graph, leading to improved performance in GAD tasks. GNN-based approaches to GAD typically involve learning representations for normal graph data or normal graph elements, followed by subsequent processing for anomaly detection. One common approach in learning representations for GAD is through the use of autoencoders. For instance, GCNAE \cite{kipf2016variational}, DOMINANT \cite{ding2019deep}, and GUIDE \cite{yuan2021higher} utilize GNNs to construct autoencoders and use reconstruction errors to calculate anomaly scores for nodes. Graph autoencoders have also been employed in the context of self-supervised learning and contrastive learning for GAD, as demonstrated in \cite{zheng2021generative, xu2022contrastive}.
Another approach is to directly output representations using neural networks. For instance, OC-GNN \cite{wang2021one} embeds graph data into a vector space and forms a hyperplane using normal data, and NetWalk \cite{yu2018netwalk} encodes graph data into a latent space using node embedding. In particular, the OC-GNN method is very similar to OCSVM, though it differs in its utilization of GNN to capture the structural information of the graph. 
However, a common limitation of existing methods that employ GNNs is their lack of interpretability. This makes it challenging to provide clear explanations for the learned representation used in decision making, and thus it is difficult to understand why specific graphs or graph elements are labeled as anomalous. In contrast, our GRAM model takes an interpretable approach that focuses on attributing outliers to explainable elements within the graph data.

In recent years, there has been a growing interest in developing interpretable methods to address the lack of interpretability in deep neural networks. Many approaches have been proposed, including the gradient-based saliency maps \cite{simonyan2013deep}, Local Interpretable Model-Agnostic Explanations (LIME) \cite{ribeiro2016should}, Excitation Backpropagation (EB) \cite{zhang2018top}, Class Activation Mapping (CAM) \cite{zhou2016learning}, and gradient-weighted CAM (Grad-CAM) \cite{selvaraju2017grad}. These methods are primarily designed for convolutional neural networks (CNNs) and aim to identify the important substructures or regions within the input data that contribute to the network's decision-making process. Accordingly, to address the interpretability limitations of deep neural networks in anomaly detection, several works have proposed methods that leverage the visual interpretation of these networks \cite{liu2020towards, salehi2021arae, jiang2023interpretability}. For instance, \cite{liu2020towards} proposed a variational autoencoder-based visual interpretation generation method which generates heatmaps for localized anomalies in images. \cite{jiang2023interpretability} proposed an anomaly detection method specifically designed for images of industrial products based on Grad-CAM attention maps, which not only utilizes attention maps in the test phase but also incorporates them into the loss function. However, these methods are not applicable for GAD tasks.

In the graph domain, there have also been recent advancements in interpretability for GNNs. \cite{pope2019explainability} proposed explainability methods for GNNs corresponding to Euclidean methods including contrastive gradient-based saliency maps, EB, CAM, Grad-CAM, and contrastive EB. Similarly, \cite{kasanishi2021edge} also extended CNN interpretability methods, such as LIME, Gradient-Based Saliency Maps, and Grad-CAM, to GNNs, with an aim to identify the edges within the input graph that are influential in the decision-making process of the GNN. However, these methods are designed for explaining the decision making process of graph classification tasks, and extending them to GAD poses a challenge. Although the idea of calculating gradient attention maps similar to a graph Grad-CAM, it is important to note that we compute and utilize gradients in a different way from \cite{pope2019explainability, kasanishi2021edge}. Furthermore, GRAM is proposed for the specific task of GAD, thus distinguishing it from the aforementioned works which focus on explaining the behavior of GNNs.

{In more recent works, \cite{miao2022interpretable} used a stochastic attention mechanism to determine the importance of different parts of the graph. \cite{chen2022learning} enhanced the GNN interpretability by aligning the model more closely with the underlying causal mechanisms of the data. These works also discussed how interpretability contributes to generalizability of GNNs. \cite{liu2024towards} developed a GAD model that not only identifies anomalies but also explains them through influential subgraphs. While \cite{liu2024towards} also incorporate interpretability, their methodology differs significantly from ours.}

\section{Proposed Method}\label{Proposed_method}

For an undirected graph $\gG$ with $N$ nodes, we denote its adjacency matrix by $\rmA \in \R^{N \times N}$ and its node feature matrix by $\rmX \in \R^{N \times M}$ where each row of $\rmX$ represents an $M$-dimensional node feature. Suppose that a GNN model takes $\rmX$ as input and produces $J$-dimensional output features as
\begin{equation}\label{eq:gnn_x_a}
\mathbf{H} = \text{GNN} \left ( \mathbf{X},\mathbf{A} \right ) . 
\end{equation}
The output $\rmH \in \R^{N \times J}$ is then post-processed to produce an $I$-dimensional vector $\rvz \in \R^I$. For instance, in our applications, $\rvz$ can be computed either from the latent distribution of a variational autoencoder or the output of a regression model. Our GRAM method leverages the gradients of $\rvz$ with respect to $\rmH$. Specifically, we define the gradient attention coefficients as follows:
\begin{equation}\label{eq:alpha_der_z_H}
\boldsymbol{\alpha}_{i} := \frac{1}{N} \sum_{n=1}^{N} \frac{\partial z_i}{\partial \rvh_n} \in \R^J, ~ i = 1, \cdots, I,
\end{equation}
where $z_i \in \R$ represents the $i$-th element of $\mathbf{z}$ and $\rvh_n \in \R^J$ represents (the transpose of) the $n$-th row of $\rmH$, corresponding to the $n$-th node. 
Next, we calculate the node-level anomaly scores for individual nodes as follows:
\begin{equation}
\label{eq:s}
\mathbf{s} \equiv [s_n]_{n=1}^N := \left[ \sum_{i=1}^I \phi (\boldsymbol{\alpha}_{i}^\T \rvh_n) \right]_{n=1}^N,
\end{equation}
where $\phi$ refers to the nonlinear activation function used for producing $\rvz$. The graph-level anomaly score is then given by
\begin{equation}
\label{eq:score}
\operatorname{score} := \operatorname{GlobalAddPooling}(\rvs) \equiv \sum_{n=1}^{N} s_n.
\end{equation}
These anomaly scores are then thresholded to determine whether individual nodes or the entire graph are classified as anomalous. The rationale behind thresholding the anomaly scores is as follows: the underlying GNN model for producing $\rvz$ is trained on normal graph data, which means that it is easier for normal nodes or graphs to explain $\rvz$ than anomalous nodes or graphs. Consequently, when examining the attention maps, anomalies are expected to exhibit significantly different scores. This will be further validated in Section \ref{Analys_GRAM_VGAE}.

GRAM is flexible and widely applicable since it does not specify the architecture of the GNN or the map from $\rmH$ to $\rvz$. In our study, we mainly focus on two types of application scenarios, in which we use the variational autoencoder model and the regression model, respectively. We present these two scenarios in detail in the following subsections.

\subsection{GRAM for anomalous graphs}
\label{GRAM_VGAE}

\begin{figure}[t]
  \centering
  \subfloat[]{\includegraphics[width=\textwidth]{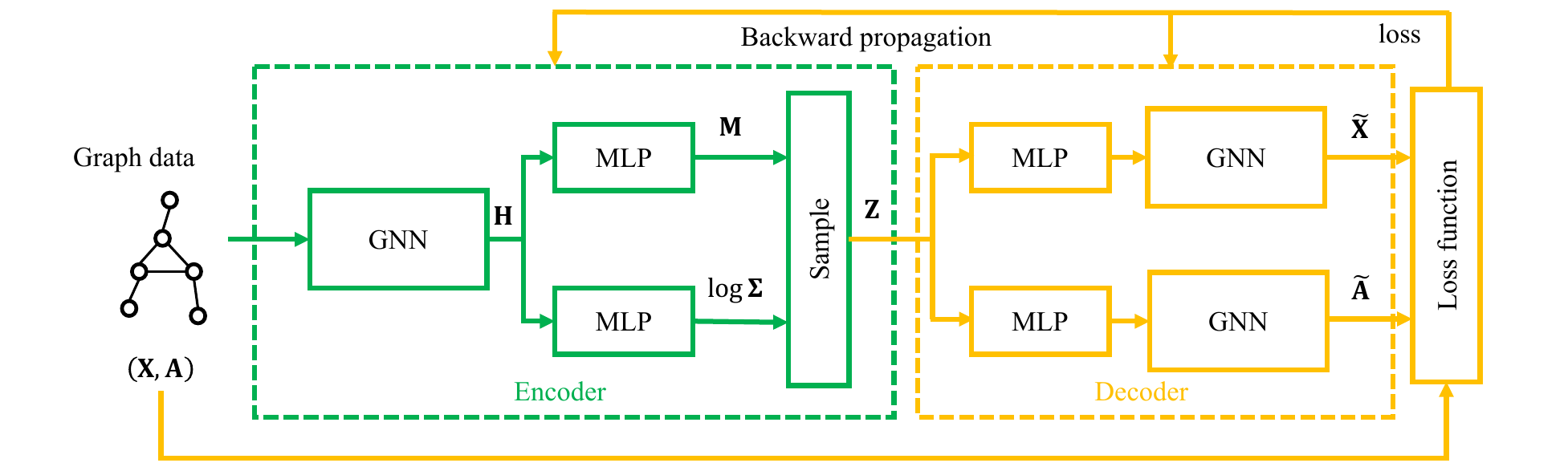}%
  \label{train}}
  \hfil
  \subfloat[]{\includegraphics[width=\textwidth]{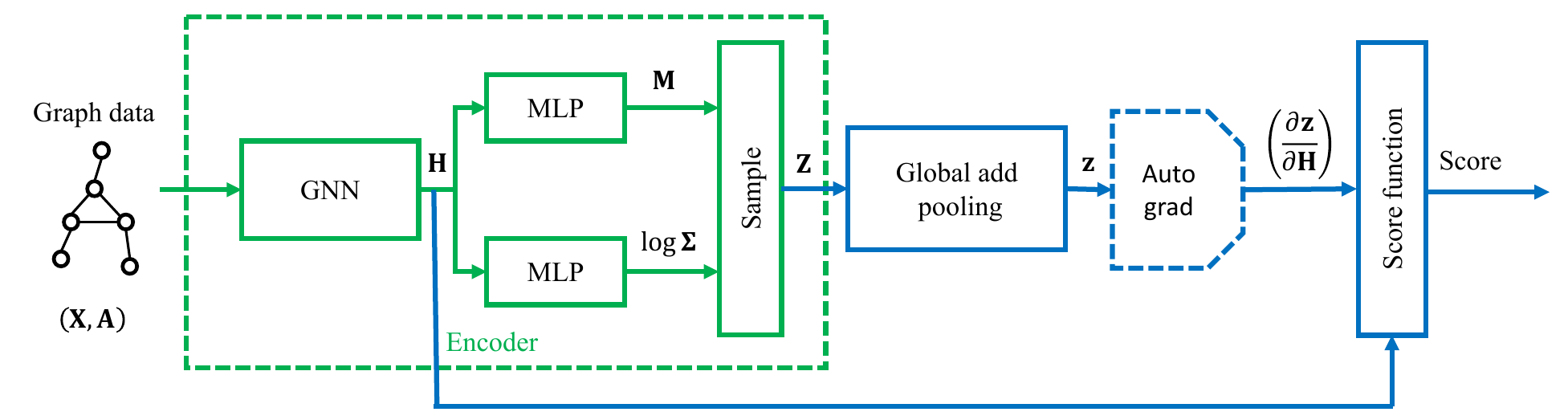}%
  \label{test}}
  \caption{The framework of the GRAM method based on the VGAE model. (a) Training phase: The VGAE model is trained according to the reconstruction error. (b) Testing phase: The encoder is employed to compute the anomaly score.}
  \label{Famework_GRAM_VGAE}
\end{figure}

In the first scenario, the dataset comprises a category of normal graphs and a category of anomalous graphs. Our goal is to accurately distinguish between normal and anomalous graphs. To address this graph-level task, we employ a variational graph autoencoder (VGAE) model similar to \cite{kipf2016variational}, but with a different architecture, to extract graph-level features and utilize the GRAM method for anomaly detection. The architectures of our VGAE during the training and testing phases of GRAM are illustrated in Fig.~\ref{Famework_GRAM_VGAE}. 

In the training phase, the VGAE is trained to minimize the unsupervised graph reconstruction error and the KL-loss. The encoder consists of a GNN and two multilayer perceptron (MLP) networks. The GNN is used to extract features from the input graph data consisting of the pair $\left ( \mathbf{X},\mathbf{A} \right )$ and generate the embedding $\mathbf{H}$. Then, the MLP networks generate the mean $\rmM$ and logarithmic standard deviation $\log\boldsymbol{\Sigma}$ for the latent variable $\mathbf{Z}$. Subsequently, the decoder consists of two MLPs and two GNNs, which decode and reconstruct the node features $\mathbf{X}$ and the adjacency matrix $\mathbf{A}$, respectively, resulting in $\tilde{\mathbf{X}}$ and $\tilde{\mathbf{A}}$. The reconstruction error is calculated by comparing the reconstructed $\tilde{\mathbf{X}}$ and $\tilde{\mathbf{A}}$ with the original input $\mathbf{X}$ and $\mathbf{A}$. The total loss for the VGAE is thus
\begin{equation}
\label{eq:loss_vage}
\beta \left \|  \mathbf{X} - \tilde{\mathbf{X}} \right \| _{\rm F}^2 + \left ( 1 - \beta \right ) \left \|  \mathbf{A} - \tilde{\mathbf{A}} \right \| _{\rm F}^2 + \text{KL-loss},
\end{equation}
where $\beta$ is a hyperparameter, $\left \| \cdot \right \| _{\rm F}$ is the Frobenius norm, and the $\text{KL-loss}$ is computed by 
\begin{equation}
\label{eq:kl_loss}
\begin{aligned}
\text{KL-loss}  &= -\frac{1}{2N} (\mathbf{1}_N)^\T \Big[ (\mathbf{1}_N) (\mathbf{1}_M)^\T + 2 \log \boldsymbol{\Sigma} \\
& \qquad -\rmM \odot \rmM   - \exp({ 2\log \boldsymbol{\Sigma}}) \Big] (\mathbf{1}_M).
\end{aligned}
\end{equation}
Here, $\mathbf{1}_N$ and $\mathbf{1}_M$ represent column vectors of all ones, with dimensions $N$ and $M$ respectively. Moreover, $\odot$ represents pointwise multiplication. 

In the testing phase, the trained encoder is utilized to produce the vector $\rvz$ for each test graph. Then, the latent vector $\mathbf{z}$ is used to calculate the anomaly score according to \eqref{eq:alpha_der_z_H}--\eqref{eq:score}.

\subsection{GRAM for anomalous nodes}\label{GRAM_RM}

\begin{figure}[t]
\centerline{\includegraphics[width=\textwidth]{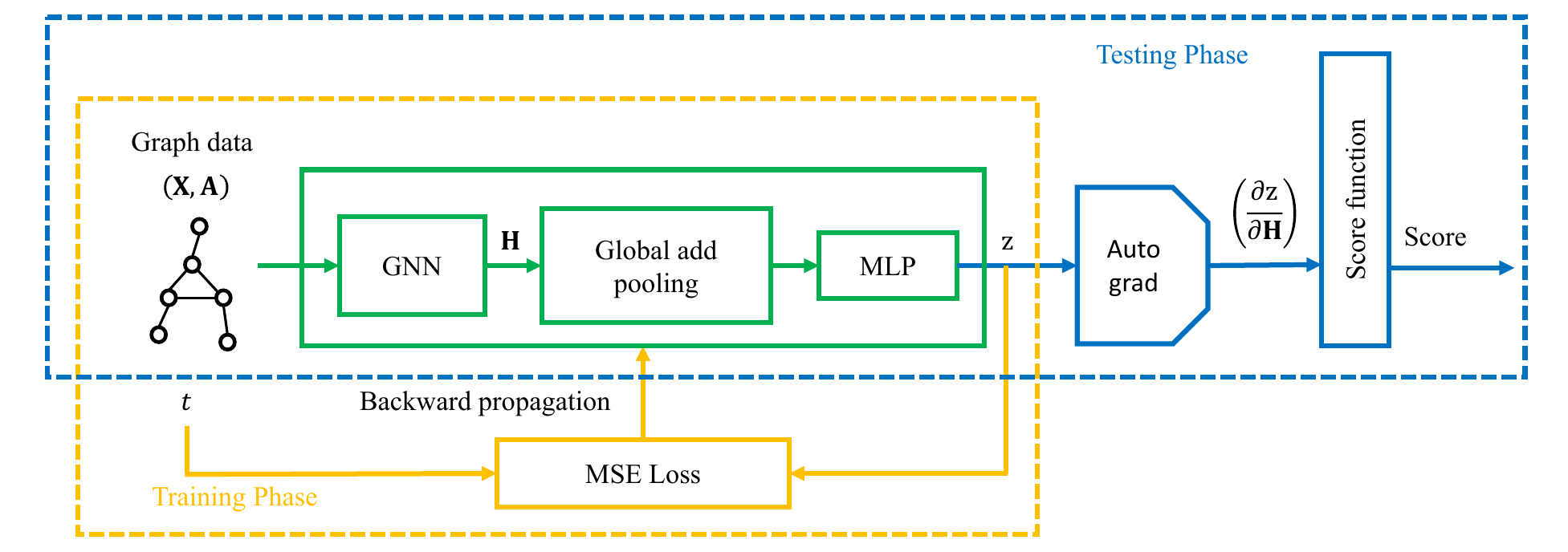}}
\caption{The framework of the GRAM method based on the regression model. Training phase: the regression model is trained using the MSE loss. Testing phase: the trained model is employed to produce the anomaly score.}
\label{model_RM}
\end{figure}

In the second scenario, our objective is to identify specific anomalous nodes within graphs. In this particular case, the dataset consists of normal graphs where each graph is associated with a graph-level target. The graph-level target represents a summarized property or value for the entire graph. We utilize a GNN regression model trained using the pair of the graph and the target, and incorporate the GRAM method to facilitate the localization of anomalous nodes within the test data, which consists of anomalous graphs that produce unexpected target values. 

Fig.~\ref{model_RM} illustrates the framework of GRAM in this scenario. In the training phase, a GNN is employed to extract node-level feature representations $\mathbf{H}$. Subsequently, a global pooling operation followed by an MLP is utilized to produce a prediction $z \in \R$ for the target value $t \in \R$. The mean-squared error (MSE) for the differences between $z$ and $t$ over all training examples is then used as the loss function for training the regression model. 

During the testing phase, the trained model is employed to make predictions for graph data in the test set while producing $\rmH$ and $z$. In this case, $z$ is a scalar, corresponding to $I=1$ in \eqref{eq:alpha_der_z_H}. We then employ \eqref{eq:alpha_der_z_H} and \eqref{eq:s} to calculate the anomaly score for each node of each graph. This facilitates the identification of anomalous nodes. 

\section{Analysis on a Simple Scenario}\label{Analys_GRAM_VGAE}
To gain insights into the regime and validate the effectiveness of GRAM, we conduct the following theoretical analysis. We consider a synthetic graph dataset which consists of two distinct types of graphs: tree-structured graphs and double ring-structured graphs, where each graph has $N$ nodes. According to our proposed method described in Section \ref{Proposed_method}, we use GRAM based on the VGAE model to address this task. To facilitate theoretical analysis, we assume very simple neural network structures which we discuss in detail later.

\begin{figure}[t]
\centerline{\includegraphics[width=0.68\columnwidth]{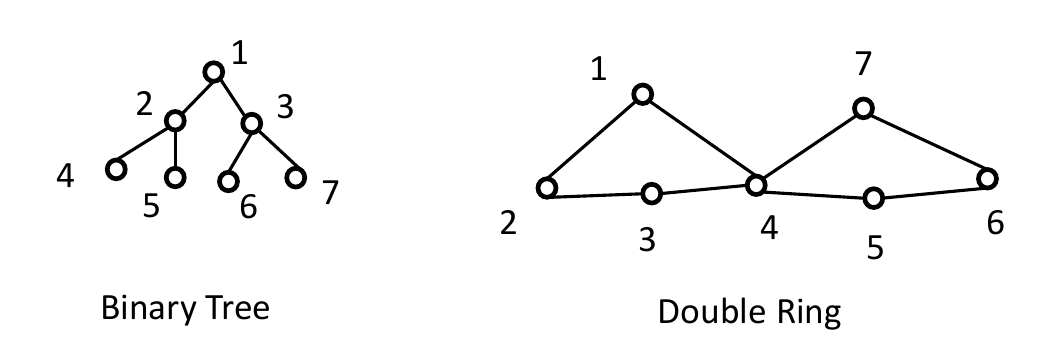}}
\caption{Examples of a binary tree graph and a double ring graph.}
\label{example_synthetic_data}
\end{figure}

In Fig.~\ref{example_synthetic_data}, we show two examples of graph data that fall in the above two categories. The left is a binary tree graph, and the right is a double ring graph. Both graphs consist of $7$ nodes but are of clearly different structures.

For simplicity, throughout this paper, we employ the graph convolutional network (GCN) \cite{kipf2016semi} as the underlying architecture for our GNN, which enables efficient feature propagation along graph edges. Specifically, we consider a GCN with $L=4$ layers, where the message passing operation in the $l$-th layer of GCN is represented as 
\begin{equation}
\mathbf{H}^{(l+1)} = \mathbf{\hat{D}}^{-\frac{1}{2}}\mathbf{\hat{A}}\mathbf{\hat{D}}^{-\frac{1}{2}}\mathbf{H}^{(l)}\mathbf{W}^{(l)}, ~l=0,1,2,3.
\end{equation}
Here, $ \mathbf{\hat{A}} = \mathbf{A} + \mathbf{I} $ denotes the adjacency matrix with inserted self-loops, $\mathbf{\hat{D}}$ refers to its corresponding diagonal degree matrix, and $\mathbf{W}^{(l)}$ represents the learnable parameters of the $l$-th layer. In addition, $\mathbf{H}^{(0)} = \mathbf{X}$ is the input feature and the output feature $\mathbf{H} = \mathbf{H}^{(L)}$. To facilitate simple calculation in our analysis, we assume no nonlinear activations between GCN layers. 
Nevertheless, we employ the rectified linear unit (ReLU), which has been shown to facilitate effective learning in various graph deep learning tasks, as the nonlinear activation function for the subsequent MLPs.
Specifically, let the MLP networks consist of two linear layers with a ReLU activation in the middle. It generates the mean $\rmM$ and logarithmic standard deviation $\log \boldsymbol{\Sigma}$ as follows:
\begin{align}
\rmM &= \text{MLP}_1(\mathbf{H}) = \text{Linear}(\text{ReLU}(\text{Linear}(\mathbf{H}))), \label{eq:mean_std} \\
\log \boldsymbol{\Sigma} &= \text{MLP}_2(\mathbf{H}) = \text{Linear}(\text{ReLU}(\text{Linear}(\mathbf{H}))). \label{eq:mean_std2}
\end{align}
We denote $\mathbf{W}_{\mu 1}$ and $\mathbf{W}_{\mu 2}$ as the weight matrices in the Linear layers in \eqref{eq:mean_std}, and $\mathbf{b}_{\mu 1}$ and $\mathbf{b}_{\mu 2}$ the bias terms. Similarly, we denote $\mathbf{W}_{\sigma 1}$ and $\mathbf{W}_{\sigma 2}$ as the weight matrices of the Linear layers in \eqref{eq:mean_std2}, and $\mathbf{b}_{\sigma 1}$ and $\mathbf{b}_{\sigma 2}$ the bias terms. With the mean and logarithmic standard deviation, the latent code is then sampled according to
\begin{equation}
\label{eq:latent_z}
\mathbf{Z} = \rmM + 
\rmE \odot \exp(\log \boldsymbol{\Sigma}),
\end{equation}
where $\rmE$ is a white noise whose entries follow the standard normal distribution independently.
The latent vector $\mathbf{z}$ is obtained from $\rmZ$ by performing global add pooling.
Therefore, the gradient of $\mathbf{z}$ with respect to the embedding $\mathbf{H}$ is
\begin{equation}\label{eq:partial_z_partial_H_chain}
\frac{\partial z_i}{\partial \mathbf{H}} = \frac{\partial z_i}{\partial \mathbf{Z}}  \frac{\partial \mathbf{Z}}{\partial \rmM}  \frac{\partial \rmM}{\partial \mathbf{H}} + \frac{\partial z_i}{\partial \mathbf{Z}}  \frac{\partial \mathbf{Z}}{\partial \log \boldsymbol{\Sigma}}  \frac{\partial \log \boldsymbol{\Sigma}}{\partial \mathbf{H}} + \frac{\partial z_i}{\partial \mathbf{Z}}  \frac{\partial \mathbf{Z}}{\partial \rmE}  \frac{\partial \rmE}{\partial \mathbf{H}},
\end{equation}
where, to simplify the representation, we abuse the notation and use $\rmH$, $\rmZ$, $\rmM$, $\boldsymbol{\Sigma}$, $\rmE$ to represent their vectorized versions, respectively.

We proceed to calculate the terms in \eqref{eq:partial_z_partial_H_chain}. First, since $\rvz$ is the sum of row vectors in $\mathbf{Z}$, 
\begin{equation}\label{eq:subs_first}
\frac{\partial z_i}{\partial \mathbf{Z}} = \mathbf{1}_N \mathbf{e}_i^\T,
\end{equation}
where $\mathbf{e}_i$ is the one-hot unit vector whose $i$-th element is $1$. Next, according to \eqref{eq:latent_z}, the partial derivative of $\mathbf{Z}$ with respect to $\rmM$ is
\begin{equation}
\frac{\partial \mathbf{Z}}{\partial \rmM} = \frac{\partial}{\partial \rmM} (\rmM + 
\rmE \odot \exp\{\log \boldsymbol{\Sigma}\}) = \mathbf{I},
\end{equation}
and the partial derivative of $\mathbf{Z}$ with respect to $\log \boldsymbol{\Sigma}$ is
\begin{equation}
\begin{aligned}
\frac{\partial \mathbf{Z}}{\partial \log \boldsymbol{\Sigma}} &= \frac{\partial}{\partial \log \boldsymbol{\Sigma}} (\rmM + 
\rmE \odot \exp\{\log \boldsymbol{\Sigma}\}) \\
&=  (\rmE \odot \exp\{\log \boldsymbol{\Sigma}\}).
\end{aligned}
\end{equation}
Further, according to \eqref{eq:mean_std} and \eqref{eq:mean_std2}, the partial derivative of $\rmM$ with respect to $\mathbf{H}$ is given by
\begin{equation}
\label{eq:mu_h}
\begin{aligned}
\frac{\partial \rmM}{\partial \mathbf{H}} &=  \frac{\partial}{\partial \mathbf{H}}(\text{Linear}(\text{ReLU}(\text{Linear}(\mathbf{H})))) \\
&= \frac{\partial}{\partial \mathbf{H}}( \text{ReLU}(\mathbf{H} \mathbf{W}_{\mu 1}  \oplus  \mathbf{b}_{\mu 1} ) \mathbf{W}_{\mu 2} \oplus \mathbf{b}_{\mu 2})  \\
&= \mathbf{W}_{\mu 2}^\T \text{ReLU}^\prime( \mathbf{H} \mathbf{W}_{\mu 1}  \oplus \mathbf{b}_{\mu 1}) \frac{\partial (\mathbf{H} \mathbf{W}_{\mu 1}  \oplus \mathbf{b}_{\mu 1})}{\partial \mathbf{H}} \\
&= \mathbf{W}_{\mu 2}^\T \text{ReLU}^\prime(\mathbf{H} \mathbf{W}_{\mu 1}  \oplus \mathbf{b}_{\mu 1}) \mathbf{W}_{\mu 1}^\T,
\end{aligned}
\end{equation}
where $\oplus$ denotes the addition of a matrix and a vector, where pointwise addition is performed after the vector is broadcasted to the same size as the matrix. Similarly, the partial derivative of $\log \boldsymbol{\Sigma}$ with respect to $\mathbf{H}$ is
\begin{equation}
\label{eq:sigma_h}
\begin{aligned}
\frac{\partial \log \boldsymbol{\Sigma}}{\partial \mathbf{H}} &= \frac{\partial}{\partial \mathbf{H}}(\text{Linear}(\text{ReLU}(\text{Linear}(\mathbf{H})))) \\
&= \frac{\partial}{\partial \mathbf{H}}( \text{ReLU}(\mathbf{H} \mathbf{W}_{\sigma 1} \oplus \mathbf{b}_{\sigma 1}) \mathbf{W}_{\sigma 2} \oplus \mathbf{b}_{\sigma 2}) \\
&= \mathbf{W}_{\sigma 2}^\T \text{ReLU}^\prime(\mathbf{H} \mathbf{W}_{\sigma 1}  \oplus \mathbf{b}_{\sigma 1}) \frac{\partial (\mathbf{H} \mathbf{W}_{\sigma 1} \oplus \mathbf{b}_{\sigma 1})}{\partial \mathbf{H}} \\
&= \mathbf{W}_{\sigma 2}^\T \text{ReLU}^\prime(\mathbf{H} \mathbf{W}_{\sigma 1}  \oplus \mathbf{b}_{\sigma 1}) \mathbf{W}_{\sigma 1}^\T.
\end{aligned}
\end{equation}
In addition,
\begin{equation}\label{eq:subs_last}
    \frac{\partial \rmE}{\partial \mathbf{H}} = 0.
\end{equation}
Finally, replacing the terms in \eqref{eq:partial_z_partial_H_chain} with \eqref{eq:subs_first}--\eqref{eq:subs_last} yields the gradient of $z_i$ with respect to $\mathbf{H}$ as follows:
\begin{equation}\label{eq:partial_z_partial_h}
\begin{aligned}
\frac{\partial z_i}{\partial \mathbf{H}} &= \frac{\partial z_i}{\partial \mathbf{Z}}  \frac{\partial \rmM}{\partial \mathbf{H}} +  \frac{\partial z_i}{\partial \mathbf{Z}}  (\rmE \odot \exp\{\log \boldsymbol{\Sigma}\})  \frac{\partial \log \boldsymbol{\Sigma}}{\partial \mathbf{H}}\\
&= (\mathbf{1}_N) (\mathbf{e}_i)^\T \Big[\mathbf{W}_{\mu 2}^\T \text{ReLU}^\prime( \mathbf{H}\mathbf{W}_{\mu 1} \oplus \mathbf{b}_{\mu 1}) \mathbf{W}_{\mu 1}^\T \\
& \quad + (\rmE \odot \exp\{\log \boldsymbol{\Sigma}\})  \mathbf{W}_{\sigma 2}^\T \\
& \quad \quad \text{ReLU}^\prime(\mathbf{H}\mathbf{W}_{\sigma 1}  \oplus \mathbf{b}_{\sigma 1}) \mathbf{W}_{\sigma 1}^\T \Big].
\end{aligned}
\end{equation}

Applying \eqref{eq:partial_z_partial_h} to \eqref{eq:alpha_der_z_H} facilitates our calculation of the gradient attention coefficients $\boldsymbol{\alpha}_i$. For the synthetic graph data we consider, we use the adjacency matrix of each graph as its node feature matrix. That is,
\begin{equation}
\label{eq:A}
\mathbf{X} = \mathbf{A} = \begin{bmatrix}
  a_{11}& \dots  & a_{1N}\\
 \vdots  & \ddots & \vdots\\
 a_{N1} & \dots & a_{nN}
\end{bmatrix}.
\end{equation}
In addition, we write out
\begin{equation}
\label{eq:A_hat_D}
\mathbf{\hat{A}} = \begin{bmatrix}
  \hat{a}_{11}& \dots  & \hat{a}_{1N}\\
 \vdots  & \vdots & \vdots\\
 \hat{a}_{N1} & \dots & \hat{a}_{nN}
\end{bmatrix}
\quad \text{and} \quad
\mathbf{\hat{D}} 
= \begin{bmatrix}
  d_{1}& \dots  & 0\\
 \vdots  & \vdots & \vdots\\
 0 & \dots & d_{N}
\end{bmatrix}.
\end{equation}
To facilitate our analysis, we assume that in both the GCN layers and the MLP layers, the weight matrices are set as the identity matrix and the bias terms are set as zero vectors. In this case, the arguments of ReLU in \eqref{eq:partial_z_partial_h} have nonnegative elements and thus making ReLU$'$ equal to $1$.
In addition, we denote
\begin{equation}
\label{eq:A_norm}
\mathbf{\hat{A}_{\text{norm}}} := \mathbf{\hat{D}}^{-\frac{1}{2}}\mathbf{\hat{A}}\mathbf{\hat{D}}^{-\frac{1}{2}}.
\end{equation}
According to our assumptions,
\begin{equation}
\label{eq:h}
\rmH^{(l)} = (\mathbf{\hat{A}_{\text{norm}}})^l \mathbf{A},~l=1,\dots,4.
\end{equation}
The $(n,j)$-th element of $\mathbf{H} = \rmH^{(4)} \in \R^{N \times N}$ ($N = I = J$ following our assumption) is thus calculated as follows: 
\begin{equation}
\label{eq:h_matrix}
\begin{aligned}
h_{nj} &= 
 \frac{1}{\sqrt{d_{n}} }\sum\limits_{k_1=1}^{N}\sum\limits_{k_2=1}^{N}\sum\limits_{k_3=1}^{N} \sum\limits_{k_4=1}^{N} \\
&\qquad \qquad \frac{\hat{a}_{nk_1}\hat{a}_{k_1k_2}\hat{a}_{k_2k_3}\hat{a}_{k_3k_4}a_{k_4 j}}{d_{k_1}d_{k_2}d_{k_3}\sqrt{d_{k_4}}}.
\end{aligned}
\end{equation}
Moreover, \eqref{eq:alpha_der_z_H} simplifies as follows:
\begin{equation}
\label{eq:alpha_matrix}
\begin{aligned}
\boldsymbol{\alpha}_{i}^{\T} & = \frac{1}{N} \mathbf{1}_N^\T  \Big[  \mathbf{1}_N\mathbf{e}_i^\T \\
& \qquad + \mathbf{1}_N\mathbf{e}_i^\T\rmE \odot \exp\{\log \boldsymbol{\Sigma}\}  \Big],
\end{aligned}
\end{equation}
where, $\rmE$ follows the centered Gaussian distribution $\gN(0, \epsilon \rmI)$ and therefore $\boldsymbol{\alpha}_i$ is also random with 
\begin{equation}
\label{eq:alpha_matrix_value}
\boldsymbol{\alpha}_{i} = \begin{bmatrix}
  \gN \left ( 0,  \epsilon \exp \left ( h_{i1}  \right ) \right ) \\  
  \vdots \\
  \gN \left ( 1,  \epsilon \exp \left ( h_{ii}  \right ) \right ) \\
  \vdots \\
  \gN \left ( 0,  \epsilon \exp \left ( h_{iN}  \right ) \right )
\end{bmatrix},
\end{equation}
where for simplicity we replace the random variables directly with the respective Gaussian distributions. Set $\boldsymbol{\alpha} = \sum\limits_{i=1}^{I} \boldsymbol{\alpha}_{i}$. We simplify \eqref{eq:s} as 
\begin{equation}
\label{eq:s_matrix}
 \mathbf{s} = \left[ \sum_{i=1}^I \boldsymbol{\alpha}_{i}^\T \rvh_n \right]_{n=1}^N = \mathbf{H}\boldsymbol{\alpha}.
\end{equation}
Combining \eqref{eq:alpha_matrix_value} and \eqref{eq:s_matrix} yields
\begin{equation}
\label{eq:s_matrix_value}
\mathbf{s} = \begin{bmatrix}
  \gN \left ( \sum\limits_{j=1}^{N} h_{1j} , \epsilon \sqrt{\sum\limits_{j=1}^{N} \sum\limits_{n=1}^{N} h_{1j}^2 \exp \left ( 2h_{nj}  \right ) }\right )    \\
  \vdots  \\
\gN \left ( \sum\limits_{j=1}^{N} h_{Nj} , \epsilon \sqrt{\sum\limits_{j=1}^{N} \sum\limits_{n=1}^{N} h_{Nj}^2 \exp \left ( 2h_{nj}  \right ) }\right )
\end{bmatrix}.
\end{equation}
Further, for the graph-level scores,
\begin{equation}
\label{eq:score_matrix_value}
\begin{aligned} 
\operatorname{score} & = \sum_{n=1}^{N} s_n = \mathbf{1}_N^{\T} \mathbf{H}\boldsymbol{\alpha}\\
& = \gN \Bigg ( \sum_{i=1}^{N}\sum_{j=1}^{N} h_{ij} , \\
& \quad \quad \quad \epsilon \sqrt{\sum_{i=1}^{N} \left(\sum_{j=1}^{N} h_{ij}  \right )^2 \sum_{n=1}^{N} \exp \left ( 2h_{ni}  \right ) }\Bigg ). 
\end{aligned}
\end{equation}

To apply the above analysis to our synthetic data, we calculate the node-level scores and graph-level scores for the two graphs in Fig.~\ref{example_synthetic_data} according to \eqref{eq:s_matrix_value} and \eqref{eq:score_matrix_value}, and examine their distinguishability. We show the results corresponding to $\epsilon = 0.1$ in Table~\ref{tab:score_syntheticdata}. 

\begin{table}[htbp]\scriptsize
\caption{Node-level and graph-level scores obtained from the GRAM method for the two graphs in Fig.~\ref{example_synthetic_data}}
\label{tab:score_syntheticdata}
\begin{center}
\renewcommand\arraystretch{1.25}
\begin{tabular}{|c|c|c|}
\hline
             & Binary tree & Double ring  \\
\hline
Node 1        & $\gN(1.94, 0.30)$                & $\gN(2.23, 0.36)$                 \\
\hline
Node 2        & $\gN(2.23, 0.37) $               & $\gN(2.24, 0.36)$               \\
  \hline
Node 3        & $\gN(2.23, 0.37)$                & $\gN(2.23, 0.36)$                 \\
\hline
Node 4        & $\gN(1.56, 0.28)$               & $\gN(2.91, 0.45)$                \\
\hline
Node 5        & $\gN(1.56, 0.28)$               & $\gN(2.23, 0.36)$                 \\
\hline
Node 6        & $\gN(1.56, 0.28)$               & $\gN(2.24, 0.36)$               \\
  \hline
Node 7        & $\gN(1.56, 0.28) $              & $\gN(2.23, 0.36) $                \\
\hline
score & $\gN(12.65, 1.74)$ & $\gN(16.32, 2.34)$ \\
\hline
\end{tabular}
\end{center}
\end{table}

First, we observe a clear separation between the graph-level scores of the two graphs, indicating that GRAM successfully captures the global differences between the graphs, which aligns with our underlying rationale. Additionally, in the case of the double ring graph, Node $4$ exhibits a higher score compared to the other nodes. This observation is significant because Node $4$ serves as the intersection point between the two rings in the graph, making it a crucial component that explains the structural dissimilarity. Consequently, the graph-level scores effectively differentiate between the two graphs, while the node-level scores provide insights into the specific nodes contributing to the dissimilarity.

The presented results showcase the effectiveness and interpretability of the anomaly scores generated by GRAM in a straightforward scenario. It is worth noting that as the neural network architecture becomes more intricate, GRAM has the capability to learn more sophisticated representations and effectively handle complex datasets. These findings lay the foundation for further exploration in a real experimental setting, which will be discussed in the following section.

\section{Experiments}\label{sec:experiments}
\subsection{GRAM for Graph-Level Anomaly Detection}\label{subsec:Ex_VGAE}

\begin{table*}[htbp]\scriptsize
\caption{Distribution of the training and test sets for various datasets}
\begin{center}
\renewcommand\arraystretch{1.25}
\begin{tabularx}{\textwidth}{|c|X|X|X|X|X|X|X|X|X|X|}
\hline
       & MUT AG   & NCI1    & PROT EINS & PTC    & IMDB-B    & RDT-B    & IMDB-M   &  RDT-M5K  & SYN data  \\
\hline
Training      & $110$                       & $1857$                      & $600$                          & $177$  & $450$                      & $900$                      & $450$                       &  $900$    &  $1800$   \\
\hline
Testing             & $30$                        & $400$                       & $126$                           & $30$      & $100$                        & $200$                       & $100$                        &  $200$   &    $400$    \\
\hline
\end{tabularx}
\label{tab:training_test_sets}
\end{center}
\end{table*}

In our first experiment, we use GRAM with an underlying VGAE model for the graph-level GAD problem.
\subsubsection{Datasets}\label{subsubsec:Ex_VGAE_Data}
We evaluate the performance of GRAM on eight real-world graph datasets, as well as the synthetic dataset constructed according to Section \ref{Analys_GRAM_VGAE}. The real-world datasets consist of graphs belonging to multiple classes. In our experimental setup, we designate one class as normal samples, while the remaining classes are considered as abnormal samples. 
Only the normal examples are available for training. The test set is constructed by sampling an equal number of normal samples and anomalous samples. 

Next, we describe the real-world datasets as follows.

\begin{itemize}
\item MUTAG \cite{debnath1991structure}: MUTAG is a dataset used for graph classification tasks. It contains chemical compounds labeled as either mutagenic or non-mutagenic based on their structure.
\item NCI1 \cite{wale2008comparison}: NCI1 is a dataset used for graph classification in the field of bioinformatics. It contains chemical compounds labeled according to their activity against non-small cell lung cancer.
\item PROTEINS \cite{borgwardt2005protein}: PROTEINS is a dataset used for protein structure classification. It consists of tertiary protein structures labeled with their fold classification.
\item PTC \cite{helma2001predictive}: PTC is a dataset used for predicting the carcinogenicity of organic compounds. It contains chemical compounds labeled as either carcinogenic or non-carcinogenic based on their structural features.
\item IMDB-B \cite{yanardag2015deep}: IMDB-B is a binary sentiment analysis dataset derived from the IMDB movie reviews. It consists of movie reviews labeled as positive or negative based on the sentiment expressed in the text. 
\item RDT-B \cite{yanardag2015deep}: RDT-B is a dataset used for spatiotemporal pattern recognition tasks. The data represent reaction-diffusion processes labeled with different patterns.
\item IMDB-M \cite{yanardag2015deep}: IMDB-M, also known as the IMDB Multiclass dataset, is a sentiment analysis dataset similar to IMDB-B. However, instead of binary labels, it contains multiclass sentiment labels representing different categories such as positive, negative, and neutral.
\item RDT-M5K \cite{yanardag2015deep}: RDT-M5K is a sentiment classification dataset derived from Reddit comments. Each graph represents a network of people, where the nodes represent individuals.
\end{itemize}

Furthermore, {we also consider a synthetic dataset (SYN data)} consisting of random tree-structured graphs as normal samples and double ring-structured graphs as anomalous samples. For all the above datasets, we present the number of graphs in the training set and test set in Table~\ref{tab:training_test_sets}.

\subsubsection{Baseline methods}\label{subsubsec:Ex_VGAE_Baseline}
We compare the GRAM method with the following baseline methods.
\begin{itemize}
\item GCNAE \cite{kipf2016variational}: GCNAE is an autoencoder framework that utilizes GCNs as both the encoder and decoder. It takes into account the graph structure and node attributes as input. The encoder learns node embeddings by aggregating neighbor information, while the decoder reconstructs node attributes using another GCN on the node embeddings and graph structure. The abnormal scores are determined according to the reconstruction error of the decoder.
\item DOMINANT \cite{ding2019deep}: DOMINANT, similar to GCNAE, also combines GCNs and autoencoders. It employs a GCN as the encoder and another GCN as the decoder to reconstruct node attributes. Additionally, it utilizes another GCN as the structural decoder and reconstructs the graph's adjacency matrix according to the similarity between decoded node features. The abnormal scores are determined by combining the reconstruction errors from both decoders.
\item GAAN \cite{chen2020generative}: GAAN is a GAN-based GAD method. It utilizes an MLP-based generator to generate fake graphs and an MLP-based encoder to encode graph information. A discriminator is trained to distinguish between connected nodes from real and fake graphs. The abnormal score is obtained by considering the node reconstruction error and the confidence in identifying real nodes.
\item CONAD \cite{xu2022contrastive}: CONAD is a GAD method that leverages graph augmentation and contrastive learning techniques to incorporate prior knowledge of outlier nodes. Augmented graphs are generated to impose the presence of outlier nodes. Siamese GNN encoders are used to encode the graphs, and contrastive loss is employed to optimize the encoder. The abnormal scores are obtained using two different decoders, similar to the DOMINANT method.
\item OC-GNN \cite{wang2021one}: OC-GNN method leverages a multilayer GNN model to process graph data and learn a representation space. The method incorporates a self-supervised learning loss function that aims to maximize the density of normal samples while minimizing the density of abnormal samples. Consequently, normal samples should cluster together in the representation space, forming close-knit clusters, while abnormal samples should be relatively isolated. Anomaly detection is done by mapping test samples to the representation space and evaluating their similarity to the clusters of normal samples.
\end{itemize}

\subsubsection{Experimental settings}\label{subsubsec:Ex_VGAE_set}
In the encoder of the VGAE, we cascade $4$ GCN layers as our GNN. Between the GCN layers, we use the Gaussian Error Linear Unit (GELU) \cite{hendrycks2016gaussian} as the nonlinear activation function. We also apply the Dropout operation. The subsequent two MLP layers are both linear layers with the GELU activation function. In the decoder, the same architectures of the MLPs and the GNNs are employed. We use the Adam optimizer for training. The hyperparameters used in GRAM include the output dimension $J$ of each node after extracting feature information by GNN, the dimension $I$ of the latent vector $\rvz$, the parameter $\beta$  of the loss function \eqref{eq:loss_vage}, the dropout rate, and the learning rate. The GNN layers have an identical number of neurons, namely $J$; and the MLP layers have an identical number of neurons, namely $I$. The specific numerics are as shown in Table~\ref{tab:parameters}. For the baseline methods, we use the public code available at \url{https://github.com/pygod-team/pygod/} but keep the GNN structures to be the same as the GNNs used in our VGAE. All the experiments were performed on a server with Intel(R) Xeon(R) Gold 6226R CPU @ 2.90GHz with NVIDIA GeForce RTX 3090 GPUs, and each experiment uses a single GPU.

\begin{table*}[htbp]\scriptsize
\caption{Hyperparameters for graph-level anomaly detection datasets}
\begin{center}
\renewcommand\arraystretch{1.25}
\begin{tabularx}{\textwidth}{|c|X|X|X|X|X|X|X|X|X|X|}
\hline
       & MUT AG   & NCI1    & PROT EINS & PTC    & IMDB-B   & RDT-B    & IMDB-M   &  RDT-M5K  & SYN data  \\
\hline
$J$                 & $128$                       & $128$                      & $128$                          & $128$          & $64$                  & $128$                      & $128$                       &  $128$    &  $128$   \\
\hline
$I$            & $64$                        & $64$                       & $64$                           & $64$         & $32$                               & $64$              & $64$                        &  $16$   &    $16$    \\
\hline
 $\beta$          & $0.25$                      & $0.25$                     & $0.3$                          & $0.25$            & $0$                    & $0.25$                     & $0$                         &    $0.25$  &  $0.25$    \\
\hline
Dropout             & $0$                          & $0$                         & $0$                            & $0$          & $0$                                  & $0$                         & $0$                        &  $0.3$    &   $0.3$  \\
\hline
Learning Rate             & $0.0007$                  & $0.0005$                 & $0.0005$                     & $0.0005$        & $0.0005$                 & $0.0005$                 & $0.0005 $                 &  $0.0005 $  & $0.0005 $ \\
\hline
\end{tabularx}
\label{tab:parameters}
\end{center}
\end{table*}

\subsubsection{Results}\label{subsubsec:Ex_VGAE_Results}
We obtain a graph-level anomaly score from each method and evaluate it using both the AUC (Area under the Receiver Operating Characteristic Curve) score and the AP (Average Precision) score. Each result is an average from three random initializations of the neural networks, reported together with the standard deviation. We present the results of AUC and AP scores in Table~\ref{tab:AUC} and Table~\ref{tab:AP}, respectively. In both tables, we use blue bold font to highlight the best performing method for each dataset, while the second best performing method is highlighted in green bold font. 

\begin{table*}[htbp]\tiny
\caption{AUC results for graph-level anomaly detection (numbers in percentage)}
\setlength\tabcolsep{2pt}
\begin{center}
\renewcommand\arraystretch{1.25}
\begin{tabular}{|c|c|c|c|c|c|c|c|c|c|}
\hline
 & MUTAG   & NCI1    & PROTEINS & PTC       & IMDB-B     & RDT-B    & IMDB-M   &  RDT-M5K  &  SYN data\\
\hline

CONAD    & $85.61 _{\pm 1.2} $ & $\bf{\textcolor{green}{65.43 _{\pm 0.1} }}$ & $74.96 _{\pm 0.0} $  & $62.22 _{\pm 0.0 }$  & $\bf{\textcolor{green}{68.36_{\pm 0.9 }}}$ & $94.30 _{\pm 0.2 }$ & $55.34_{\pm 0.0  }$  & $90.20 _{\pm 0.4 }$ & $\bf{\textcolor{green}{97.72 _{\pm 0.3 }}}$\\
\hline
DOMINANT & $85.61 _{\pm 1.2 }$ & $65.36 _{\pm 0.1 }$ & $\bf{\textcolor{green}{75.01 _{\pm 0.1 }}}$  & $62.22_{\pm 0.0 }$  & $67.40 _{\pm 2.7 }$ & $94.72 _{\pm 0.2 }$ & $51.71 _{\pm 2.0 }$ & $89.96 _{\pm 0.1 }$ & $\bf{\textcolor{green}{97.72 _{\pm 0.3 }}}$\\
\hline
GAAN    & $84.45_{\pm 1.6 }$ & $64.46_{\pm 0.0 }$ & $72.41_{\pm 0.0 }$  & $60.89_{\pm 0.0 }$   & $63.91_{\pm 1.1 }$ & $\bf{\textcolor{blue}{97.12_{\pm 0.0 }}}$ & $54.12_{\pm 0.0 }$  & $88.26 _{\pm 0.0 }$ & $93.27 _{\pm 0.3 }$\\
\hline
GCNAE   & $79.97 _{\pm 0.8 }$ & $62.27 _{\pm 0.2 }$ & $72.30 _{\pm 0.0 }$  & $\bf{\textcolor{green}{66.67 _{\pm 0.0 }}}$ & $56.99 _{\pm 8.8 }$  & $91.28_{\pm 0.1 }$ & $\bf{\textcolor{green}{59.35 _{\pm 0.0 }}}$  & $\bf{\textcolor{blue}{95.06 _{\pm 0.1 }}}$ & $91.85 _{\pm 5.1 }$\\
\hline
OC-GNN  & $\bf{\textcolor{blue}{88.59 _{\pm 0.3 }}}$ & $59.30 _{\pm 6.6 }$ & $74.54 _{\pm 1.1 }$ & $58.22 _{\pm 0.5 }$ & $61.97 _{\pm 7.2 }$ & $94.16 _{\pm 1.0 }$ & $54.71 _{\pm 6.2 }$ & $73.73 _{\pm 16.0 }$ & $95.06 _{\pm 1.9 }$\\
\hline
GRAM   & $\bf{\textcolor{green}{86.99 _{\pm 1.7 }}}$ & $\bf{\textcolor{blue}{68.31 _{\pm 0.3 }}}$ & $\bf{\textcolor{blue}{76.66 _{\pm 0.1 }}}$  & $\bf{\textcolor{blue}{76.15 _{\pm 2.9 }}}$ & $\bf{\textcolor{blue}{70.76 _{\pm 1.4 }}}$ & $\bf{\textcolor{green}{94.95 _{\pm 0.8 }}}$ & $\bf{\textcolor{blue}{61.76 _{\pm 1.0 }}}$  & $\bf{\textcolor{green}{94.26 _{\pm 2.1 }}}$ & $\bf{\textcolor{blue}{99.49 _{\pm 0.2}}}$\\
\hline
\end{tabular}
\label{tab:AUC}
\end{center}
\end{table*}

\begin{table*}[htbp]\tiny
\caption{AP results for graph-level anomaly detection (numbers in percentage)}
\setlength\tabcolsep{2pt}
\begin{center}
\renewcommand\arraystretch{1.25}
\begin{tabular}{|c|c|c|c|c|c|c|c|c|c|}
\hline
       & MUTAG   & NCI1    & PROTEINS & PTC    & IMDB-B    & RDT-B    & IMDB-M   &  RDT-M5K &  SYN data\\
\hline
CONAD    & $86.26 _{\pm 0.8 }$ & $\bf{\textcolor{green}{69.29 _{\pm 0.0 }}}$ & $76.05 _{\pm 0.0 }$  & $55.46 _{\pm 0.0 }$  & $\bf{\textcolor{green}{66.42 _{\pm 3.1 }}}$ & $95.75 _{\pm 0.1 }$ & $55.27 _{\pm 3.0 }$ & $88.75 _{\pm 0.4 }$  & $\bf{\textcolor{green}{97.96 _{\pm 0.2 }}}$\\
\hline
DOMINANT & $86.26 _{\pm 0.8 }$ & $69.24 _{\pm 0.0 }$ & $\bf{\textcolor{green}{76.10 _{\pm 0.1 }}}$  & $55.49 _{\pm 0.0 }$ & $65.65 _{\pm 3.9 }$  & $96.01 _{\pm 0.1 }$ & $51.52 _{\pm 1.4 }$ & $88.64 _{\pm 0.1 }$  & $\bf{\textcolor{green}{97.96 _{\pm 0.2 }}}$\\
\hline
GAAN     & $84.09 _{\pm 0.8 }$ & $66.93 _{\pm 0.0 }$ & $73.18 _{\pm 0.0 }$  & $53.87 _{\pm 0.0 }$  & $63.37 _{\pm 0.8 }$ & $\bf{\textcolor{blue}{97.44 _{\pm 0.0 }}}$ & $53.22 _{\pm 4.5 }$ & $87.09 _{\pm 0.0 }$ & $92.12 _{\pm 0.4 }$ \\
\hline
GCNAE    & $78.47 _{\pm 1.9 }$ & $68.13 _{\pm 0.0 }$ & $73.93 _{\pm 0.0 }$  & $58.99 _{\pm 0.0 }$  & $61.02 _{\pm 9.8 }$ & $93.20 _{\pm 0.1 }$ & $55.62 _{\pm 0.0 }$ & $\bf{\textcolor{green}{90.46 _{\pm 0.1 }}}$ & $90.30 _{\pm 4.9 }$ \\
\hline
OC-GNN  & $\bf{\textcolor{blue}{89.28 _{\pm 0.1 }}}$ & $58.20 _{\pm 3.6 }$ & $73.24 _{\pm 1.2 }$ & $\bf{\textcolor{green}{67.24 _{\pm 0.5 }}}$ & $63.04 _{\pm 8.4 }$ & $90.05 _{\pm 0.3 }$ & $\bf{\textcolor{green}{56.29 _{\pm 7.5 }}}$ & $75.74 _{\pm 18.1 }$ & $91.83 _{\pm 4.1 }$ \\
\hline
GRAM  & $\bf{\textcolor{green}{87.22 _{\pm 2.4}} }$ & $\bf{\textcolor{blue}{70.25 _{\pm 0.1 }}}$ & $\bf{\textcolor{blue}{76.96 _{\pm 0.1 }}}$  & $\bf{\textcolor{blue}{70.67 _{\pm 5.6 }}}$  & $\bf{\textcolor{blue}{69.57 _{\pm 1.7 }}}$ & $\bf{\textcolor{green}{96.10 _{\pm 0.5 }}}$ & $\bf{\textcolor{blue}{59.68 _{\pm 3.3 }}}$ & $\bf{\textcolor{blue}{95.81 _{\pm 0.7 }}}$ & $\bf{\textcolor{blue}{99.26 _{\pm 0.7 }}}$\\
\hline
\end{tabular}
\label{tab:AP}
\end{center}
\end{table*}

From the results presented in Table~\ref{tab:AUC}, it is evident that the GRAM method consistently achieves excellent AUC performance on all nine datasets. It outperforms other methods on a majority of the datasets, ranking first on six datasets and ranking second on the remaining three datasets. Similarly, Table~\ref{tab:AP} shows that the GRAM method achieves the best AP performance on seven datasets and it achieves the second highest AP scores on the remaining two datasets. We also observe that for all the baseline methods, there are cases where the performance is significantly worse than our GRAM method, indicating their inconsistency. These results indicate that GRAM consistently demonstrates competitive performance in addressing the graph-level GAD problem.

\begin{table}[t!]\tiny
\caption{{AUC results for graph-level anomaly detection (numbers in percentage), with backbone GNN layers changed to GraphConv}}
\setlength\tabcolsep{2pt}
\begin{center}
\renewcommand\arraystretch{1.25}
\begin{tabular}{|l*{9}{|c}|}
\hline
& MUTAG & NCI1 & PROTEINS & PTC & IMDB-B & RDT-B & IMDB-M & RDT-M5K & SYN data\\
\hline
CONAD & $84.88_{\pm0.0}$ & $65.51_{\pm0.1}$ & $\bf{\textcolor{green}{74.90_{\pm0.1}}}$ & $62.07_{\pm0.3}$ & $52.39_{\pm1.7}$ & $87.85_{\pm9.5}$ & $57.67_{\pm0.3}$ & $85.46_{\pm0.5}$ & $\bf{\textcolor{green}{97.92_{\pm0.0}}}$ \\
\hline
DOMINANT & $85.03_{\pm0.3}$ & $65.44_{\pm0.0}$ & $70.33_{\pm0.5}$ & $62.52_{\pm0.5}$ & $\bf{\textcolor{green}{64.71_{\pm1.0}}}$ & $\bf{\textcolor{blue}{99.37_{\pm0.2}}}$ & $60.22_{\pm0.2}$ & $\bf{\textcolor{green}{86.89_{\pm0.8}}}$ & $\bf{\textcolor{green}{97.92_{\pm0.0}}}$ \\
\hline
GCNAE & $62.37_{\pm1.6}$ & $\bf{\textcolor{blue}{68.10_{\pm2.5}}}$ & $72.26_{\pm1.2}$ & $\bf{\textcolor{green}{63.70_{\pm3.2}}}$ & $53.50_{\pm2.5}$ & $93.95_{\pm0.1}$ & $\bf{\textcolor{blue}{63.21_{\pm0.6}}}$ & $72.62_{\pm0.8}$ & $95.75_{\pm0.7}$ \\
\hline
OC-GNN & $\bf{\textcolor{blue}{92.59_{\pm0.7}}}$ & $63.33_{\pm0.3}$ & $54.96_{\pm3.6}$ & $54.96_{\pm3.6}$ & $55.80_{\pm0.6}$ & $89.32_{\pm0.9}$ & $56.70_{\pm4.9}$ & $81.21_{\pm1.4}$ & $94.02_{\pm0.1}$ \\
\hline
GRAM & $\bf{\textcolor{green}{87.11_{\pm0.8}}}$ & $\bf{\textcolor{green}{67.20_{\pm0.3}}}$ & $\bf{\textcolor{blue}{75.23_{\pm0.1}}}$ & $\bf{\textcolor{blue}{74.52_{\pm0.5}}}$ & $\bf{\textcolor{blue}{70.05_{\pm0.9}}}$ & $\bf{\textcolor{green}{98.29_{\pm0.2}}}$ & $\bf{\textcolor{green}{61.17_{\pm0.1}}}$ & $\bf{\textcolor{blue}{88.60_{\pm2.3}}}$ & $\bf{\textcolor{blue}{98.07_{\pm0.0}}}$ \\
\hline
\end{tabular}
\label{tab:Graphconv_AUC}
\end{center}
\end{table}

\begin{table}[t!]\tiny
\caption{{AP results for graph-level anomaly detection (numbers in percentage), with backbone GNN layers changed to GraphConv}}
\setlength\tabcolsep{2pt}
\begin{center}
\renewcommand\arraystretch{1.25}
\begin{tabular}{|l*{9}{|c}|}
\hline
& MUTAG & NCI1 & PROTEINS & PTC & IMDB-B & RDT-B & IMDB-M & RDT-M5K & SYN data \\
\hline
CONAD & $85.80_{\pm0.0}$ & $\bf{\textcolor{green}{69.27_{\pm0.0}}}$ & $\bf{\textcolor{green}{75.91_{\pm0.2}}}$ & $55.28_{\pm0.2}$ & $55.58_{\pm2.0}$ & $89.95_{\pm9.2}$ & $57.83_{\pm0.1}$ & $81.98_{\pm2.0}$ & $\bf{\textcolor{blue}{98.09_{\pm0.0}}}$ \\
\hline
DOMINANT & $85.96_{\pm0.3}$ & $69.26_{\pm0.0}$ & $69.86_{\pm0.3}$ & $55.80_{\pm1.0}$ & $\bf{\textcolor{green}{60.55_{\pm0.4}}}$ & $\bf{\textcolor{blue}{99.45_{\pm0.1}}}$ & $59.07_{\pm0.1}$ & $\bf{\textcolor{blue}{84.66_{\pm1.0}}}$ & $\bf{\textcolor{blue}{98.09_{\pm0.0}}}$ \\
\hline
GCNAE & $62.29_{\pm0.9}$ & $\bf{\textcolor{green}{69.01_{\pm3.4}}}$ & $73.16_{\pm0.9}$ & $\bf{\textcolor{green}{61.49_{\pm4.2}}}$ & $54.96_{\pm2.5}$ & $95.54_{\pm0.1}$ & $\bf{\textcolor{blue}{80.43_{\pm0.7}}}$ & $66.32_{\pm0.6}$ & $96.38_{\pm0.6}$ \\
\hline
OC-GNN & $\bf{\textcolor{blue}{93.34_{\pm1.0}}}$ & $61.15_{\pm0.2}$ & $73.48_{\pm2.9}$ & $57.00_{\pm2.9}$ & $55.09_{\pm0.2}$ & $80.22_{\pm1.9}$ & $58.12_{\pm2.7}$ & $77.57_{\pm1.0}$ & $89.46_{\pm0.2}$ \\
\hline
GRAM & $\bf{\textcolor{green}{89.75_{\pm0.4}}}$ & $\bf{\textcolor{blue}{69.61_{\pm1.2}}}$ & $\bf{\textcolor{blue}{77.93_{\pm0.3}}}$ & $\bf{\textcolor{blue}{79.44_{\pm0.7}}}$ & $\bf{\textcolor{blue}{66.51_{\pm1.6}}}$ & $\bf{\textcolor{green}{98.46_{\pm0.2}}}$ & $\bf{\textcolor{green}{61.83_{\pm0.1}}}$ & $\bf{\textcolor{green}{83.86_{\pm2.7}}}$ & $\bf{\textcolor{green}{98.07_{\pm0.0}}}$ \\
\hline
\end{tabular}
\label{tab:Graphconv_AP}
\end{center}
\end{table}

{Furthermore, to showcase the flexibility of GRAM, we also conduct experiments using a different GNN layer architecture. In addition to the above results of experiments where the backbone GNNs are taken to be GCN, we also conduct the same experiments using GraphConv~\cite{morris2019weisfeiler}, with the results shown in Tables \ref{tab:Graphconv_AUC} and \ref{tab:Graphconv_AP}. We remark that GAAN employs MLP and is excluded in the new result. The results presented in Tables \ref{tab:Graphconv_AUC} and \ref{tab:Graphconv_AP} demonstrate the outstanding performance of GRAM across all datasets, surpassing other methods in most cases. These findings indicate that GRAM is not limited by a fixed GNN layer architecture but can instead flexibly adapt to different architectures.}

\subsubsection{Interpretative performance}\label{subsubsec:Inter_result}
\begin{figure}[t]
\centerline{\includegraphics[width=\columnwidth]{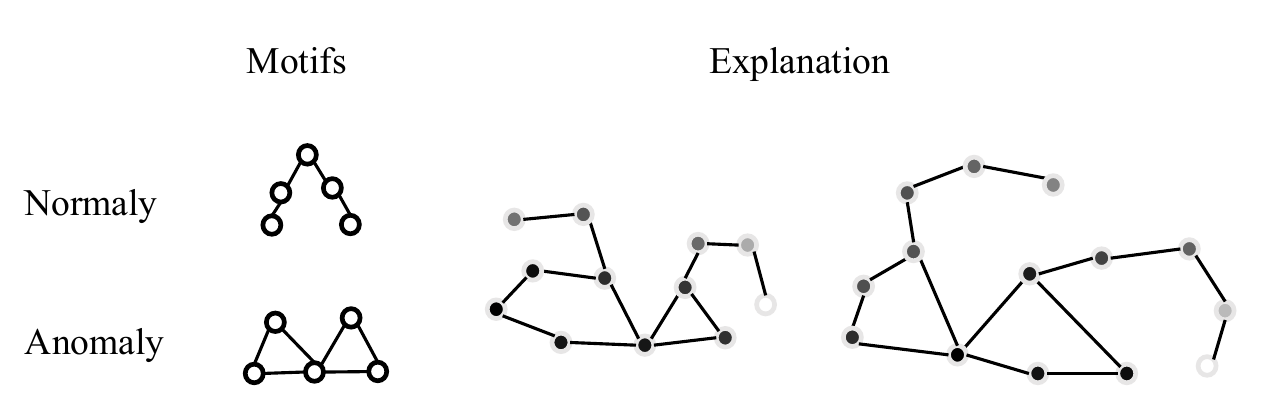}}
\caption{{Visualization of anomaly scores for sampled SYN data. These scores are mapped to the grayscale intensities of the graph nodes for clarity.}}
\label{SYN_plot}
\end{figure}

{GRAM involves training GNN on normal graph data and then utilizing gradient-based attention maps to obtain interpretability. When examining the gradient attention maps of test data, anomalous samples exhibit significantly different scores, thus effectively explaining the decision process of the GRAM method. 
To see the decision process clearly, we visualize node-level scores on the synthetic data named SYN, described in Section~\ref{subsubsec:Ex_VGAE_Data}, to demonstrate the interpretability of GRAM. In the visualization of Fig.~\ref{SYN_plot}, abnormal samples contain double ring motifs. We illustrate the gradient-based anomaly score as the grayscale intensity of nodes. We observe that the nodes within the double ring have higher anomaly scores.}

{To demonstrate the interpretability performance of the GRAM on real-word datasets, we utilized three datasets containing ground-truth explanations: MNIST-0, MNIST-1, and MUTAG, following the same setting of \cite{liu2024towards}. MNIST-0 and MNIST-1, constructed from the MNIST-75sp superpixel dataset~\cite{knyazev2019understanding}, where nodes and edges that have non-zero pixel values within the images serve as the ground-truth explanations. In the MUTAG dataset, following \cite{luo2020parameterized}, the -NO2 and -NH2 patterns in anomalous molecules are identified as the ground-truth explanations.}

{To showcase the competitive interpretability performance of the GRAM method, we compared it not only with the baseline methods mentioned in Section \ref{subsubsec:Ex_VGAE_Baseline} but also with the state-of-the-art SIGNET method \cite{liu2024towards} specifically designed for GAD interpretability. For quantitative evaluation of interpretability, we adopt the same metrics as in \cite{liu2024towards, miao2022interpretable}, reporting the AUC results for node-level explanation (NX-AUC) and edge-level explanation (EX-AUC). NX-AUC is calculated based on node-level anomaly scores, while EX-AUC is calculated based on edge-level anomaly scores derived from the node-level anomaly scores of node pairs corresponding to edges. The specific results are presented in Table \ref{tab:NX_EX}. To comprehensively evaluate the performance of the SIGNET method, we report results from experiments conducted using the code provided by \cite{liu2024towards} on our machine, labeled as ``SIGNET''. Additionally, we record the scores reported in their original paper as ``SIGNET-P''. It can be observed that the interpretability results of the GRAM outperform all baseline methods, demonstrating its superior performance in interpretability.}
\begin{table}[t!]\tiny
\caption{{NX-AUC and EX-AUC on real-word datasets (numbers in percentage)}}
\begin{center}
\renewcommand\arraystretch{1.25}
\setlength\tabcolsep{5.75pt}
\begin{tabular}{|l|c|c|c|c|c|c|}
\hline
 & \multicolumn{2}{c|}{MNIST-0} & \multicolumn{2}{c|}{MNIST-1} & \multicolumn{2}{c|}{MUTAG} \\
\hline
 & NX-AUC & EX-AUC & NX-AUC & EX-AUC & NX-AUC & EX-AUC \\
\hline
CONAD & $76.97_{\pm0.43}$ & $74.08_{\pm0.17}$ & $69.15_{\pm0.59}$ & $70.43_{\pm0.72}$ & $76.20_{\pm3.44}$ & $84.75_{\pm2.81}$ \\
\hline
DOMINANT & $\bf{\textcolor{green}{77.26_{\pm0.32}}}$ & $\bf{\textcolor{green}{74.23_{\pm0.09}}}$ & $\bf{\textcolor{green}{69.36_{\pm0.66}}}$ & $70.48_{\pm0.92}$ & $52.24_{\pm2.22}$ & $70.86_{\pm3.63}$ \\
\hline
GAAN & $75.92_{\pm1.88}$ & $70.98_{\pm2.24}$ & $66.93_{\pm0.42}$ & $66.36_{\pm0.53}$ & $68.85_{\pm2.12}$ & $81.66_{\pm6.94}$ \\
\hline
GCNAE & $74.22_{\pm2.81}$ & $71.20_{\pm2.61}$ & $62.54_{\pm1.49}$ & $60.41_{\pm0.68}$ & $\bf{\textcolor{green}{82.49_{\pm2.50}}}$ & $85.24_{\pm1.42}$ \\
\hline
OC-GNN & $58.88_{\pm3.21}$ & $57.99_{\pm3.51}$ & $54.35_{\pm2.23}$ & $54.25_{\pm2.50}$ & $81.50_{\pm4.37}$ & $\bf{\textcolor{green}{85.55_{\pm5.32}}}$ \\
\hline
SIGNET & $71.54_{\pm2.04}$ & $74.03_{\pm3.42}$ & $67.71_{\pm3.39}$ & $73.28_{\pm4.81}$ & $66.61_{\pm13.10}$ & $71.11_{\pm16.04}$ \\
\hline
SIGNET-P & $70.38_{\pm5.64}$ & $72.78_{\pm7.25}$ & $68.44_{\pm3.07}$ & $\bf{\textcolor{green}{74.83_{\pm5.24}}}$ & $75.24_{\pm8.94}$ & $78.05_{\pm9.19}$ \\
\hline
GRAM & $\bf{\textcolor{blue}{84.25_{\pm0.71}}}$ & $\bf{\textcolor{blue}{85.13_{\pm0.30}}}$ & $\bf{\textcolor{blue}{83.89_{\pm1.92}}}$ & $\bf{\textcolor{blue}{89.60_{\pm0.60}}}$ & $\bf{\textcolor{blue}{86.42_{\pm9.39}}}$ & $\bf{\textcolor{blue}{87.73_{\pm9.29}}}$ \\
\hline
\end{tabular}
\label{tab:NX_EX}
\end{center}
\end{table}

\subsection{GRAM for Node-Level Anomaly Detection}\label{subsec:Ex_RM}
In our second experiment, we use GRAM with an underlying regression model to localize anomalous nodes in graphs. 
\subsubsection{Datasets}\label{subsubsec:Ex_RM_Data}
We perform experiments on a wireless network dataset taken from \cite{yang2023graph}. The dataset contains wireless communication networks, each modeled as a weighted graph where the weight is given by the communication rate. Each network in the dataset comprises six nodes, including one source node, one target node, and four relay nodes. It contains information about the locations of the nodes. It also contains a corresponding maximum throughput as a target value, which can be regarded as a function of the locations of the nodes. During normal operation, each node maintains a stable and efficient communication rate with the other nodes. We construct a training set consisting of $160,000$ networks with normal operation. To construct the anomalous data, we take $1,000$ networks independent from the training data. In each network, we randomly take a relay node and halve the communication rate between that node and other nodes. In practice, this decline in communication performance has a direct impact on the maximum throughput of the entire wireless communication network, leading to anomalies in its overall performance. Our objective is to detect this specific anomalous node.

We remark that while there are other datasets available for node-level GAD, they encounter two main challenges. First, the anomalies present in these datasets are often artificially generated without a real application motivation, and some non-graph methods have shown comparable performance to graph-based methods \cite{liu2022bond, gu2023three}. Second, these datasets do not facilitate learning multiple graphs, which is a fundamental requirement for the GRAM framework. Therefore, we exclude these datasets in our experiments.

\subsubsection{Baseline methods}\label{subsubsec:Ex_RM_Baseline}
We test the same baseline methods as reviewed in Section~\ref{subsubsec:Ex_VGAE_Baseline}. Additionally, we test the following baseline method, which was excluded in the first experiment because it requires the same number of nodes in all the graphs.
\begin{itemize}
\item GUIDE \cite{yuan2021higher}: GUIDE uses an attribute autoencoder and a structure autoencoder to reconstruct node attributes and structure information, respectively. The structural information used is a node structure vector that can encode higher-order structural information represented by the node motif degree. The reconstruction error is used to calculate the anomaly score of each node.
\end{itemize}

\subsubsection{Experimental settings}\label{subsubsec:Ex_RM_set}
In the regression model, we cascade three GraphConv layers \cite{morris2019weisfeiler} as our GNN. The GNN is followed by a global add pooling operation, and an MLP with two linear layers is utilized as the output layer. In both the GNN and the MLP, we use the GELU activation function between the layers. There are $32$ neurons in both the GraphConv layers and the MLP layers, except for the last linear layer of MLP, which only contains $1$ neuron for the scalar output. During the training process, we take a batch size of $100$ and use the Adam optimizer with a learning rate of $0.0002$. Since the baseline approaches are not regression models, we incorporate the target value as an additional dimension to the node features to facilitate fair comparison. All the experiments were performed on the same server as mentioned in Section \ref{subsubsec:Ex_VGAE_set}.

\subsubsection{Results}\label{subsubsec:Ex_RM_Results}

The evaluation metric in this part is the accuracy rate, which refers to the percentage of accurately identified anomalous nodes among the $1,000$ graphs in the test set. We present the results in Table~\ref{tab:result_wireless}.

\begin{table}[htbp]\scriptsize
\caption{Performance of various methods on the wireless network dataset (numbers in percentage)}
\begin{center}
\renewcommand\arraystretch{1.25}
\begin{tabular}{|c|c|}
\hline
       & Accuracy rate  \\
\hline
CONAD    & $23.40$        \\
\hline
DOMINANT & $23.90$        \\
\hline
GAAN     & $25.30$        \\
\hline
GCNAE    & $22.10$        \\
\hline
GUIDE    & $22.30$        \\
\hline
OC-GNN  & $\bf{\textcolor{green}{42.90}}$    \\
\hline
GRAM  & $\bf{\textcolor{blue}{61.60}}$       \\
\hline
\end{tabular}
\label{tab:result_wireless}
\end{center}
\end{table}

Analyzing the result, we observe that five of the baseline methods, namely CONAD, DOMINANT, GAAN, GCNAE, and GUIDE, do not achieve satisfactory performance. It is important to note that in these wireless network scenarios, there are only four relay nodes present. Therefore, a random guess would yield an accuracy rate of approximately $25\%$. The method OC-GNN shows some improvement compared to the above methods. Nevertheless, our GRAM method significantly outperforms all the baseline methods. This highlights the effectiveness of our proposed method in the scenario of node-level GAD.

\section{Conclusion}\label{sec:Conclusion}
In this work, we proposed the GRAM method as an interpretable approach for anomaly detection for GAD. Specifically, for datasets that consist of both normal and abnormal graph samples and the goal is to distinguish abnormal graphs, we train a VGAE model in an unsupervised manner and then use its encoder to extract graph-level features for computing the anomaly scores. For datasets consisting of graphs with graph-level labels, where the goal is to identify specific anomalous nodes in anomalous graphs, we train a GNN regression model using the labels, and then use the gradient attention as a post-processing step to accurately locate the anomalous nodes. 

We performed a theoretical analysis using synthetic data, making simplistic assumptions about the neural networks, to gain insights into the decision-making process of the GRAM method. Furthermore, we compared the performance of the GRAM method with state-of-the-art baseline methods on the respective tasks, demonstrating its competitive performance.

There are some limitations to the current work. First, GRAM relies on the assumption that the training data consists of normal graphs, ensuring that the GNN learns accurate features and the interpretation remains reliable. Second, it is designed for training on multiple graphs and may not be applicable to scenarios where anomalies are present within a single graph. In future research, we intend to enhance the GRAM method to address these limitations and adapt to the aforementioned scenarios.

\nocite*
\bibliographystyle{IEEEtran}
\bibliography{IEEEabrv, ref}

\begin{thebibliography}{10}
\providecommand{\url}[1]{#1}
\csname url@samestyle\endcsname
\providecommand{\newblock}{\relax}
\providecommand{\bibinfo}[2]{#2}
\providecommand{\BIBentrySTDinterwordspacing}{\spaceskip=0pt\relax}
\providecommand{\BIBentryALTinterwordstretchfactor}{4}
\providecommand{\BIBentryALTinterwordspacing}{\spaceskip=\fontdimen2\font plus
\BIBentryALTinterwordstretchfactor\fontdimen3\font minus \fontdimen4\font\relax}
\providecommand{\BIBforeignlanguage}[2]{{%
\expandafter\ifx\csname l@#1\endcsname\relax
\typeout{** WARNING: IEEEtran.bst: No hyphenation pattern has been}%
\typeout{** loaded for the language `#1'. Using the pattern for}%
\typeout{** the default language instead.}%
\else
\language=\csname l@#1\endcsname
\fi
#2}}
\providecommand{\BIBdecl}{\relax}
\BIBdecl

\bibitem{nguyen2020fang}
V.-H. Nguyen, K.~Sugiyama, P.~Nakov, and M.-Y. Kan, ``Fang: Leveraging social context for fake news detection using graph representation,'' in \emph{Proc. 29th ACM Int. Conf. Inf. Knowl. Manage.}, 2020, pp. 1165--1174.

\bibitem{kumar2018rev2}
S.~Kumar, B.~Hooi, D.~Makhija, M.~Kumar, C.~Faloutsos, and V.~Subrahmanian, ``Rev2: Fraudulent user prediction in rating platforms,'' in \emph{Proc. 11th ACM Int. Conf. Web Search Data Mining}, 2018, pp. 333--341.

\bibitem{miao2020attack}
K.~Miao, X.~Shi, and W.-A. Zhang, ``Attack signal estimation for intrusion detection in industrial control system,'' \emph{Comput. Secur.}, vol.~96, p. 101926, 2020.

\bibitem{wu2024beyond}
J.~Wu, R.~Hu, D.~Li, L.~Ren, Z.~Huang, and Y.~Zang, ``Beyond the individual: An improved telecom fraud detection approach based on latent synergy graph learning,'' \emph{Neural Netw.}, vol. 169, pp. 20--31, 2024.

\bibitem{zhang2022efraudcom}
G.~Zhang, Z.~Li, J.~Huang, J.~Wu, C.~Zhou, J.~Yang, and J.~Gao, ``efraudcom: An e-commerce fraud detection system via competitive graph neural networks,'' \emph{ACM Trans. Inf. Syst.}, vol.~40, no.~3, pp. 1--29, 2022.

\bibitem{pang2021deep}
G.~Pang, C.~Shen, L.~Cao, and A.~V.~D. Hengel, ``Deep learning for anomaly detection: A review,'' \emph{ACM Comput. Surv.}, vol.~54, no.~2, pp. 1--38, 2021.

\bibitem{akoglu2015graph}
L.~Akoglu, H.~Tong, and D.~Koutra, ``Graph based anomaly detection and description: a survey,'' \emph{Data Min. Knowl. Discovery.}, vol.~29, pp. 626--688, 2015.

\bibitem{jie2019block}
F.~Jie, C.~Wang, F.~Chen, L.~Li, and X.~Wu, ``Block-structured optimization for anomalous pattern detection in interdependent networks,'' in \emph{Proc. IEEE Int. Conf. Data Mining}.\hskip 1em plus 0.5em minus 0.4em\relax IEEE, 2019, pp. 1138--1143.

\bibitem{ma2021comprehensive}
X.~Ma, J.~Wu, S.~Xue, J.~Yang, C.~Zhou, Q.~Z. Sheng, H.~Xiong, and L.~Akoglu, ``A comprehensive survey on graph anomaly detection with deep learning,'' \emph{IEEE Trans. Knowl. Data Eng.}, 2021.

\bibitem{breunig2000lof}
M.~M. Breunig, H.-P. Kriegel, R.~T. Ng, and J.~Sander, ``{LOF}: identifying density-based local outliers,'' in \emph{Proc. Int. Conf. Manage. Data (SIGMOD)}, 2000, pp. 93--104.

\bibitem{scholkopf2001estimating}
B.~Sch{\"o}lkopf, R.~C. Williamson, A.~J. Smola, J.~Shawe-Taylor, and J.~C. Platt, ``Estimating the support of a high-dimensional distribution,'' \emph{Neural Comput.}, vol.~13, no.~7, pp. 1443--1471, 2001.

\bibitem{erfani2016high}
S.~M. Erfani, S.~Rajasegarar, S.~Karunasekera, and C.~Leckie, ``High-dimensional and large-scale anomaly detection using a linear one-class {SVM} with deep learning,'' \emph{Pattern Recognit.}, vol.~58, pp. 121--134, 2016.

\bibitem{zhai2016deep}
S.~Zhai, Y.~Cheng, W.~Lu, and Z.~Zhang, ``Deep structured energy based models for anomaly detection,'' in \emph{Proc. 33rd Int. Conf. Mach. Learn.}, vol.~48.\hskip 1em plus 0.5em minus 0.4em\relax PMLR, 2016, pp. 1100--1109.

\bibitem{zong2018deep}
B.~Zong, Q.~Song, M.~R. Min, W.~Cheng, C.~Lumezanu, D.~Cho, and H.~Chen, ``Deep autoencoding gaussian mixture model for unsupervised anomaly detection,'' in \emph{Int. Conf. Learn. Represent.}, 2018.

\bibitem{sabokrou2018adversarially}
M.~Sabokrou, M.~Khalooei, M.~Fathy, and E.~Adeli, ``Adversarially learned one-class classifier for novelty detection,'' in \emph{Proc. IEEE Conf. Comput. Vis. Pattern Recognit.}, 2018, pp. 3379--3388.

\bibitem{pidhorskyi2018generative}
S.~Pidhorskyi, R.~Almohsen, and G.~Doretto, ``Generative probabilistic novelty detection with adversarial autoencoders,'' in \emph{Proc. 32nd Int. Conf. Neural Informat. Process. Syst.}, 2018, pp. 6822--6833.

\bibitem{zenati2018adversarially}
H.~Zenati, M.~Romain, C.-S. Foo, B.~Lecouat, and V.~Chandrasekhar, ``Adversarially learned anomaly detection,'' in \emph{Proc. IEEE Int. Conf. Data Mining}.\hskip 1em plus 0.5em minus 0.4em\relax IEEE, 2018, pp. 727--736.

\bibitem{perera2019ocgan}
P.~Perera, R.~Nallapati, and B.~Xiang, ``{OCGAN}: One-class novelty detection using gans with constrained latent representations,'' in \emph{Proc. IEEE Conf. Comput. Vis. Pattern Recognit.}, 2019, pp. 2898--2906.

\bibitem{xie2020unsupervised}
W.~Xie, Y.~Li, J.~Lei, J.~Yang, J.~Li, X.~Jia, and Z.~Li, ``Unsupervised spectral mapping and feature selection for hyperspectral anomaly detection,'' \emph{Neural Netw.}, vol. 132, pp. 144--154, 2020.

\bibitem{lai2020robust}
C.-H. Lai, D.~Zou, and G.~Lerman, ``Robust subspace recovery layer for unsupervised anomaly detection,'' in \emph{Int. Conf. Learn. Represent.}, 2020.

\bibitem{chen2020generative}
Z.~Chen, B.~Liu, M.~Wang, P.~Dai, J.~Lv, and L.~Bo, ``Generative adversarial attributed network anomaly detection,'' in \emph{Proc. 29th ACM Int. Conf. Inf. Knowl. Manage.}, 2020, pp. 1989--1992.

\bibitem{zha2020meta}
D.~Zha, K.-H. Lai, M.~Wan, and X.~Hu, ``Meta-{AAD}: Active anomaly detection with deep reinforcement learning,'' in \emph{Proc. IEEE Int. Conf. Data Mining}.\hskip 1em plus 0.5em minus 0.4em\relax IEEE, 2020, pp. 771--780.

\bibitem{liguori2021adversarial}
A.~Liguori, G.~Manco, F.~S. Pisani, and E.~Ritacco, ``Adversarial regularized reconstruction for anomaly detection and generation,'' in \emph{Proc. IEEE Int. Conf. Data Mining}.\hskip 1em plus 0.5em minus 0.4em\relax IEEE, 2021, pp. 1204--1209.

\bibitem{chen2022utrad}
L.~Chen, Z.~You, N.~Zhang, J.~Xi, and X.~Le, ``{UTRAD}: Anomaly detection and localization with u-transformer,'' \emph{Neural Netw.}, vol. 147, pp. 53--62, 2022.

\bibitem{lai2023robust}
C.-H. Lai, D.~Zou, and G.~Lerman, ``Robust variational autoencoding with wasserstein penalty for novelty detection,'' in \emph{Proc. Int. Conf. Artif. Intell. Statist.}\hskip 1em plus 0.5em minus 0.4em\relax PMLR, 2023, pp. 3538--3567.

\bibitem{kipf2016variational}
T.~N. Kipf and M.~Welling, ``Variational graph auto-encoders,'' \emph{NIPS Workshop on Bayesian Deep Learning}, 2016.

\bibitem{ding2019deep}
K.~Ding, J.~Li, R.~Bhanushali, and H.~Liu, ``Deep anomaly detection on attributed networks,'' in \emph{Proc. SIAM Int. Conf. Data Mining}.\hskip 1em plus 0.5em minus 0.4em\relax SIAM, 2019, pp. 594--602.

\bibitem{yuan2021higher}
X.~Yuan, N.~Zhou, S.~Yu, H.~Huang, Z.~Chen, and F.~Xia, ``Higher-order structure based anomaly detection on attributed networks,'' in \emph{Proc. IEEE Int. Conf. Big Data (Big Data)}.\hskip 1em plus 0.5em minus 0.4em\relax IEEE, 2021, pp. 2691--2700.

\bibitem{zheng2021generative}
Y.~Zheng, M.~Jin, Y.~Liu, L.~Chi, K.~T. Phan, and Y.-P.~P. Chen, ``Generative and contrastive self-supervised learning for graph anomaly detection,'' \emph{IEEE Trans. Knowl. Data Eng.}, 2021.

\bibitem{xu2022contrastive}
Z.~Xu, X.~Huang, Y.~Zhao, Y.~Dong, and J.~Li, ``Contrastive attributed network anomaly detection with data augmentation,'' in \emph{Proc. 26th Pacific-Asia Conf. Knowl. Discov. Data Mining (PAKDD)}.\hskip 1em plus 0.5em minus 0.4em\relax Springer, 2022, pp. 444--457.

\bibitem{wang2021one}
X.~Wang, B.~Jin, Y.~Du, P.~Cui, Y.~Tan, and Y.~Yang, ``One-class graph neural networks for anomaly detection in attributed networks,'' \emph{Neural Comput. Appl.}, vol.~33, pp. 12\,073--12\,085, 2021.

\bibitem{yu2018netwalk}
W.~Yu, W.~Cheng, C.~C. Aggarwal, K.~Zhang, H.~Chen, and W.~Wang, ``Netwalk: A flexible deep embedding approach for anomaly detection in dynamic networks,'' in \emph{Proc. ACM SIGKDD 24th Int. Conf. Knowl. Discov. Data Mining,}, 2018, pp. 2672--2681.

\bibitem{simonyan2013deep}
K.~Simonyan, A.~Vedaldi, and A.~Zisserman, ``Deep inside convolutional networks: Visualising image classification models and saliency maps,'' \emph{arXiv:1312.6034}, 2013.

\bibitem{ribeiro2016should}
M.~T. Ribeiro, S.~Singh, and C.~Guestrin, ````{W}hy should {I} trust you?'' {E}xplaining the predictions of any classifier,'' in \emph{Proc. 22nd ACMSIGKDD Int. Conf. Knowl. Discovery Data Mining}, 2016, pp. 1135--1144.

\bibitem{zhang2018top}
J.~Zhang, S.~A. Bargal, Z.~Lin, J.~Brandt, X.~Shen, and S.~Sclaroff, ``Top-down neural attention by excitation backprop,'' \emph{Int. J. Comput. Vis.}, vol. 126, no.~10, pp. 1084--1102, 2018.

\bibitem{zhou2016learning}
B.~Zhou, A.~Khosla, A.~Lapedriza, A.~Oliva, and A.~Torralba, ``Learning deep features for discriminative localization,'' in \emph{Proc. IEEE Conf. Comput. Vis. Pattern Recognit.}, 2016, pp. 2921--2929.

\bibitem{selvaraju2017grad}
R.~R. Selvaraju, M.~Cogswell, A.~Das, R.~Vedantam, D.~Parikh, and D.~Batra, ``Grad-{CAM}: Visual explanations from deep networks via gradient-based localization,'' in \emph{Proc. IEEE Int. Conf. Comput. Vis.}, 2017, pp. 618--626.

\bibitem{liu2020towards}
W.~Liu, R.~Li, M.~Zheng, S.~Karanam, Z.~Wu, B.~Bhanu, R.~J. Radke, and O.~Camps, ``Towards visually explaining variational autoencoders,'' in \emph{Proc. IEEE/CVF Conf. Comput. Vis. Pattern Recognit.}, 2020, pp. 8642--8651.

\bibitem{salehi2021arae}
J.~Zhang, S.~A. Bargal, Z.~Lin, J.~Brandt, X.~Shen, and S.~Sclaroff, ``{ARAE}: Adversarially robust training of autoencoders improves novelty detection,'' \emph{Neural Netw.}, vol. 144, pp. 726--736, 2021.

\bibitem{jiang2023interpretability}
R.~Jiang, Y.~Xue, and D.~Zou, ``Interpretability-aware industrial anomaly detection using autoencoders,'' \emph{IEEE Access}, vol.~11, pp. 60\,490--60\,500, 2023.

\bibitem{pope2019explainability}
P.~E. Pope, S.~Kolouri, M.~Rostami, C.~E. Martin, and H.~Hoffmann, ``Explainability methods for graph convolutional neural networks,'' in \emph{Proc. IEEE Conf. Comput. Vis. Pattern Recognit.}, 2019, pp. 10\,772--10\,781.

\bibitem{kasanishi2021edge}
T.~Kasanishi, X.~Wang, and T.~Yamasaki, ``Edge-level explanations for graph neural networks by extending explainability methods for convolutional neural networks,'' in \emph{Proc. IEEE Int. Symp. Multimedia}.\hskip 1em plus 0.5em minus 0.4em\relax IEEE, 2021, pp. 249--252.

\bibitem{miao2022interpretable}
S.~Miao, M.~Liu, and P.~Li, ``Interpretable and generalizable graph learning via stochastic attention mechanism,'' in \emph{Proc. Int. Conf. Mach. Learn.}\hskip 1em plus 0.5em minus 0.4em\relax PMLR, 2022, pp. 15\,524--15\,543.

\bibitem{chen2022learning}
Y.~Chen, Y.~Zhang, Y.~Bian, H.~Yang, M.~Kaili, B.~Xie, T.~Liu, B.~Han, and J.~Cheng, ``Learning causally invariant representations for out-of-distribution generalization on graphs,'' \emph{Adv. Neural Inform. Process. Syst.}, vol.~35, pp. 22\,131--22\,148, 2022.

\bibitem{liu2024towards}
Y.~Liu, K.~Ding, Q.~Lu, F.~Li, L.~Y. Zhang, and S.~Pan, ``Towards self-interpretable graph-level anomaly detection,'' \emph{Adv. Neural Inform. Process. Syst.}, vol.~36, 2024.

\bibitem{kipf2016semi}
T.~N. Kipf and M.~Welling, ``Semi-supervised classification with graph convolutional networks,'' in \emph{Int. Conf. Learn. Represent.}, 2017.

\bibitem{debnath1991structure}
A.~K. Debnath, R.~L. Lopez~de Compadre, G.~Debnath, A.~J. Shusterman, and C.~Hansch, ``Structure-activity relationship of mutagenic aromatic and heteroaromatic nitro compounds. correlation with molecular orbital energies and hydrophobicity,'' \emph{J. Medicinal Chem.}, vol.~34, no.~2, pp. 786--797, 1991.

\bibitem{wale2008comparison}
N.~Wale, I.~A. Watson, and G.~Karypis, ``Comparison of descriptor spaces for chemical compound retrieval and classification,'' \emph{Knowl. Inf. Syst.}, vol.~14, pp. 347--375, 2008.

\bibitem{borgwardt2005protein}
K.~M. Borgwardt, C.~S. Ong, S.~Sch{\"o}nauer, S.~Vishwanathan, A.~J. Smola, and H.-P. Kriegel, ``Protein function prediction via graph kernels,'' \emph{Bioinformatics}, vol.~21, no. suppl\_1, pp. i47--i56, 2005.

\bibitem{helma2001predictive}
C.~Helma, R.~D. King, S.~Kramer, and A.~Srinivasan, ``The predictive toxicology challenge 2000--2001,'' \emph{Bioinformatics}, vol.~17, no.~1, pp. 107--108, 2001.

\bibitem{yanardag2015deep}
P.~Yanardag and S.~Vishwanathan, ``Deep graph kernels,'' in \emph{Proc. 21th ACM SIGKDD Int. Conf. Knowl. Discov. Data Mining}, 2015, pp. 1365--1374.

\bibitem{hendrycks2016gaussian}
D.~Hendrycks and K.~Gimpel, ``Gaussian error linear units ({GELU}s),'' \emph{arXiv:1606.08415}, 2016.

\bibitem{morris2019weisfeiler}
C.~Morris, M.~Ritzert, M.~Fey, W.~L. Hamilton, J.~E. Lenssen, G.~Rattan, and M.~Grohe, ``{Weisfeiler} and {Leman} go neural: Higher-order graph neural networks,'' in \emph{Proc. AAAI Conf. Artif. Intell.}, vol.~33, no.~01, 2019, pp. 4602--4609.

\bibitem{knyazev2019understanding}
B.~Knyazev, G.~W. Taylor, and M.~Amer, ``Understanding attention and generalization in graph neural networks,'' \emph{Adv. Neural Inform. Process. Syst.}, vol.~32, 2019.

\bibitem{luo2020parameterized}
D.~Luo, W.~Cheng, D.~Xu, W.~Yu, B.~Zong, H.~Chen, and X.~Zhang, ``Parameterized explainer for graph neural network,'' \emph{Adv. Neural Inform. Process. Syst.}, vol.~33, pp. 19\,620--19\,631, 2020.

\bibitem{yang2023graph}
Y.~Yang, D.~Zou, and X.~He, ``Graph neural network-based node deployment for throughput enhancement,'' \emph{IEEE Trans. Neural Netw. Learn. Syst.}, pp. 1--15, 2023.

\bibitem{liu2022bond}
K.~Liu, Y.~Dou, Y.~Zhao, X.~Ding, X.~Hu, R.~Zhang, K.~Ding, C.~Chen, H.~Peng, K.~Shu, L.~Sun, J.~Li, G.~H. Chen, Z.~Jia, and P.~S. Yu, ``{BOND}: Benchmarking unsupervised outlier node detection on static attributed graphs,'' in \emph{Proc. 36th Int. Conf. Neural Informat. Process. Syst.}, 2022.

\bibitem{gu2023three}
J.~Gu and D.~Zou, ``Three revisits to node-level graph anomaly detection: Outliers, message passing and hyperbolic neural networks,'' in \emph{Proc. The Second Learn. Graphs Conf.}, 2023.

\bibitem{zhao2021synergistic}
T.~Zhao, T.~Jiang, N.~Shah, and M.~Jiang, ``A synergistic approach for graph anomaly detection with pattern mining and feature learning,'' \emph{IEEE Trans. Neural Netw. Learn. Syst.}, vol.~33, no.~6, pp. 2393--2405, 2021.

\bibitem{wang2022making}
Y.~Wang, C.~Qin, Y.~Bai, Y.~Xu, X.~Ma, and Y.~Fu, ``Making reconstruction-based method great again for video anomaly detection,'' in \emph{Proc. IEEE Int. Conf. Data Mining}.\hskip 1em plus 0.5em minus 0.4em\relax IEEE, 2022, pp. 1215--1220.

\bibitem{ruff2018deep}
L.~Ruff, R.~Vandermeulen, N.~Goernitz, L.~Deecke, S.~A. Siddiqui, A.~Binder, E.~M{\"u}ller, and M.~Kloft, ``Deep one-class classification,'' in \emph{Proc. Int. Conf. Mach. Learn.}\hskip 1em plus 0.5em minus 0.4em\relax PMLR, 2018, pp. 4393--4402.

\bibitem{ying2019gnnexplainer}
Z.~Ying, D.~Bourgeois, J.~You, M.~Zitnik, and J.~Leskovec, ``Gnnexplainer: Generating explanations for graph neural networks,'' \emph{Adv. Neural Inform. Process. Syst.}, vol.~32, 2019.

\end{thebibliography}




\end{document}